\documentclass[10pt,journal,compsoc]{IEEEtran}

\usepackage{xcolor,soul,framed} 

\colorlet{shadecolor}{yellow}
\usepackage[pdftex]{graphicx}
\graphicspath{{../pdf/}{../jpeg/}}
\DeclareGraphicsExtensions{.pdf,.jpeg,.png}

\usepackage[cmex10]{amsmath}
\usepackage{array}
\usepackage{mdwmath}
\usepackage{mdwtab}
\usepackage{eqparbox}
\usepackage{url}
\usepackage{graphicx}
\usepackage{amsmath}
\usepackage{amsfonts}
\usepackage{algorithm}
\usepackage{algpseudocode}
\usepackage{multirow}
\usepackage{footnote}
\usepackage{bm}
\usepackage{makecell}
\usepackage{pdflscape}
\usepackage{afterpage}
\usepackage{bbding}
\usepackage{lscape}
\usepackage{booktabs}
\usepackage{tikz}
\usetikzlibrary{mindmap}  
\usepackage{bbding}
\usepackage{pgfplots}
\usepackage{caption}
\usepackage{rotating}
\usepackage{tabularray}
\usepackage{tabularx}
\usepackage{makecell}
\usepackage{ragged2e}
\usepackage{threeparttable}
\usepackage{amssymb}
\usepackage{colortbl}
\usepackage{color}

\newcolumntype{L}{>{\RaggedRight\hangafter=1\hangindent=0em}X}
\newcolumntype{C}{>{\Centering\hangafter=1\hangindent=0em}X}

\pgfplotsset{compat=1.12}
\usepackage{subcaption}
    \pgfplotsset{
        layers/my layer set/.define layer set={
            background,
            main,
            foreground
        }{ },
        set layers=my layer set,
    }

\usepackage[square,sort,comma,numbers]{natbib} 
\usepackage[top=0.6in,bottom=0.4in,left=0.4in,textwidth=7.7in]{geometry}
\usepackage[pagebackref=false,breaklinks=false,linkcolor=red,anchorcolor=black, citecolor=green,colorlinks,bookmarks=true]{hyperref}
\newcommand{\etc}{\textit{etc}. }

\newcommand{\eg}{\textit{e}.\textit{g}., }
\usepackage{amsfonts}

\usepackage{pifont}

\begin{document}





\title{ATRNet-STAR: A Large Dataset and Benchmark \protect\\ Towards Remote Sensing Object \protect\\ Recognition in the Wild}
\author{Yongxiang Liu$^{^*}$, Weijie Li, Li Liu$^{^*}$, Jie Zhou, Bowen Peng, Yafei Song, Xuying Xiong, \\Wei Yang, Tianpeng Liu, Zhen Liu, Xiang Li$^{^*}$
\thanks{
This work was supported by National Natural Science Foundation of China (NSFC) under Grant 62376283 and 62531026, the Science and Technology Innovation Program of Hunan Province under Grant 2022RC1092, and Innovation Research Foundation of National University of Defense Technology. \emph{($^{\ast}$Corresponding authors: Yongxiang Liu, Li Liu, and Xiang Li. E-mail address: lyx\_bible@sina.com, liuli\_nudt@nudt.edu.cn and lixiang01@vip.sina.com.)}
}
\thanks{The authors are with the College of Electronic Science and Technology, National University of Defense Technology, Changsha, 410073, China. }
}
\markboth{Submitted to IEEE Trans. Pattern Analysis and Machine Intelligence}%
{Li \MakeLowercase{\textit{et al.}}: SARATR-X}

\IEEEtitleabstractindextext{
\begin{center}\setcounter{figure}{0}
	\centering
	\vspace{-0.6cm}
	\includegraphics[width=0.92\textwidth]{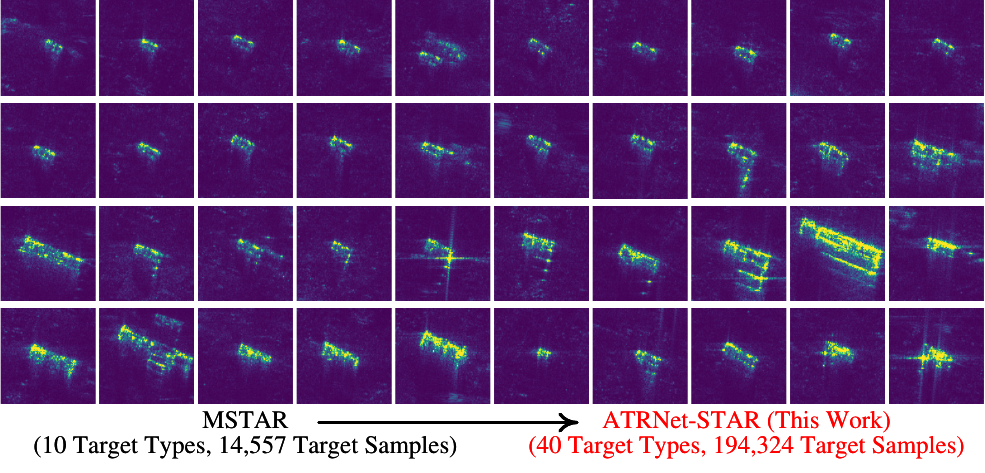}
	\vspace{-0.6cm}
	\captionof{figure}{
		\small
		\textcolor{black}{Our ATRNet-STAR dataset contains 40 distinct target types, collected with the aim of replacing the outdated though widely used MSTAR dataset and making a significant contribution to the advancement of SAR ATR research.}
	}\label{show}
\end{center}

\begin{abstract}
 \justifying 
The absence of publicly available, large-scale, high-quality datasets for Synthetic Aperture Radar Automatic Target Recognition (SAR ATR) has significantly hindered the application of rapidly advancing deep learning techniques, which hold huge potential to unlock new capabilities in this field. This is primarily because collecting large volumes of diverse target samples from SAR images is prohibitively expensive, largely due to privacy concerns, the characteristics of microwave radar imagery perception, and the need for specialized expertise in data annotation. Throughout the history of SAR ATR research, there have been only a number of small datasets, mainly including targets like ships, airplanes, buildings, \etc There is only one vehicle dataset MSTAR collected in the 1990s, which has been a valuable source for SAR ATR. To fill this gap, this paper introduces a large-scale, new dataset named ATRNet-STAR with 40 different vehicle categories collected under various realistic imaging conditions and scenes. It marks a substantial advancement in dataset scale and diversity, comprising over 190,000 well-annotated samples\textemdash $10\times$ larger than its predecessor, the famous MSTAR. Building such a large dataset is a challenging task, and the data collection scheme will be detailed. Secondly, we illustrate the value of ATRNet-STAR via extensively evaluating the performance of 15 representative methods with 7 different experimental settings on challenging classification and detection benchmarks derived from the dataset. Finally, based on our extensive experiments, we identify valuable insights for SAR ATR and discuss potential future research directions in this field. We hope that the scale, diversity, and benchmark of ATRNet-STAR can significantly facilitate the advancement of SAR ATR.

\end{abstract}

} 

\maketitle
\IEEEdisplaynontitleabstractindextext
\IEEEpeerreviewmaketitle

\IEEEraisesectionheading{\section{Introduction}
\label{Introduction}}
\begin{figure*}[!tb]
\centering
\includegraphics[width=18.0cm]{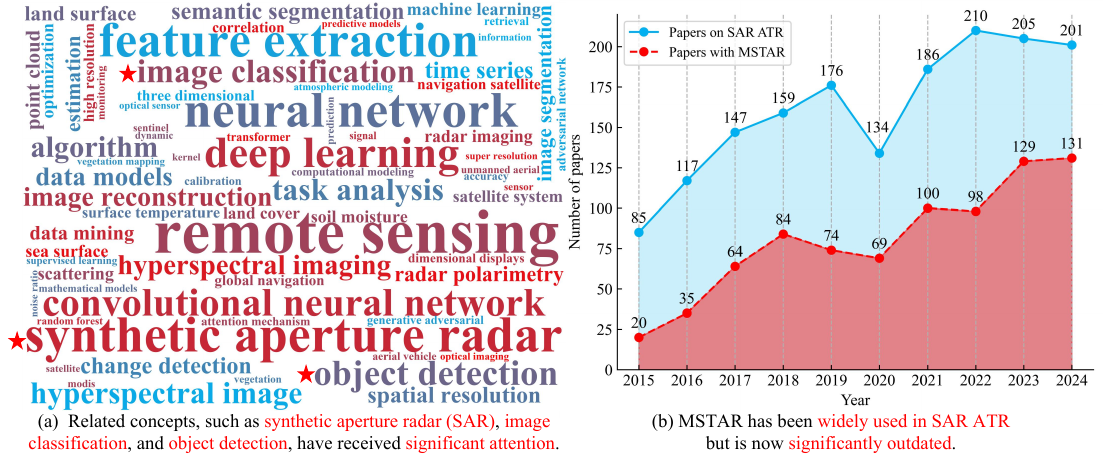}
\caption{\textbf{Motivation of our ATRNet-STAR.} 
\textbf{Subfigure (a)} depicts the most frequent keywords in 21,780 journal papers published in remote sensing (TGRS, JSTARS, GRSL, ISPRS Journal, and JAG) from 2020 to 2024. The size of each word is proportional to its frequency, highlighting that concepts such as Synthetic Aperture Radar (SAR), image classification, and object detection have garnered substantial attention.
\textbf{Subfigure (b)} focuses on the number of publications related to SAR Automatic Target Recognition (ATR) over the past five years, a cross-area of the concepts highlighted in Subfigure (a). As the pioneering dataset for SAR target classification, MSTAR has long served as the predominant benchmark due to its unique data diversity and accumulated benchmarks. 
}
\label{fig_motivation}
\end{figure*}
\IEEEPARstart{S}ynthetic Aperture Radar (SAR) imaging is capable of generating high-resolution imagery irrespective of lighting conditions and weather~\cite{xiong2024quality, huang2024generative,zhu2021meet,sun2021spaceborne}, and has become an indispensable tool for Earth observation~\cite{gagliardi2023satellite, huang2024generative,tsokas2022sar}. SAR Remote Sensing (RS) data enables analysis and recognition of objects and scenes~\cite{huang2024generative,zhu2021meet,dong2020keypoint}, which has become a valuable complement to other RS imagery~\cite{10168277,zhang2024optical, hong2024spectralgpt, hu2024conceptual}. Consequently, as a fundamental and challenging field in RS image analysis, SAR Automatic Target Recognition (ATR)~\cite{kechagias2021automatic,zhou2024diffdet4sar}, which autonomously detects and classifies objects of interest (\eg vehicles, ships, aircraft, and buildings), has become an active research area for several decades (Fig.~\ref{fig_motivation} (a)). SAR ATR has a wide range of civilian
and military applications, including global surveillance, military reconnaissance, urban management, disaster assessment, and
emergency rescue~\cite{guo2024skysense,zhangearthgpt,wang2025attributed,astruc2024anysat, pengs2mtea}. 
Despite the remarkable achievements over the past several decades in the field of SAR ATR, the accurate, robust, and efficient recognition of any target in an open world remains unresolved~\cite{kechagias2021automatic, dong2020keypoint, 11315131}.

\textbf{\textcolor{black}{Need for ATRNet -}}
The advent of big data has propelled the evolution of RS pre-training foundation models~\cite{lu2024ai, zheng2024changen2,wang2024hypersigma,gong2024crossearth} where large-scale pre-training enables efficient cross-task adaptation with minimal finetuning. However, the scarcity of large-scale standardized datasets restricts the development of generalizable data-driven SAR ATR methods, compared to the success of foundation models in other RS sensors~\cite{zhou2024towards,10768939,wu2023fully}. The data sensitivity, acquisition costs, annotation difficulty, and complexity of SAR imaging hinder the establishment of open large-scale data ecosystems: More than 50\% of SAR ATR studies still rely on the 1990s-era Moving and Stationary Target Acquisition and Recognition (MSTAR)~\cite{MSTAR} dataset (Fig.~\ref{fig_motivation} (b)). In addition, non-standardized evaluation protocols across existing benchmarks (\eg MSTAR, OpenSARShip~\cite{huang2018opensarship}, and FUSAR-Ship~\cite{hou2020fusar}) impede objective algorithm comparison. Therefore, building a large SAR ATR dataset and benchmark is necessary to unlock new model capabilities in this field. We aim for \textbf{ATRNet}\textemdash a massive, diverse, and standard SAR target dataset benchmark for modeling and recognizing target characteristics. 

\IEEEpubidadjcol

\textbf{\textcolor{black}{Need for ATRNet-STAR\footnote{STAR: Stationary Target Acquisition and Recognition. Our dataset focuses on the fine-grained single target recognition problem in SAR ATR, \emph{i.e.}, the image classification task, but we provide a horizontal bounding box label for the object detection task in different conditions.} -}}
Researchers have struggled to construct many SAR target datasets for ATR tasks. Many SAR classification datasets in Table~\ref{table:atr_dataset} have significantly improved data diversity. In particular, new SAR detection datasets~\cite{li2024sardet,wu2024fair} have emerged with 100,000 images. 
However, our previous research~\cite{li2024predicting,li2024saratrx} on SAR foundation models revealed that collecting public datasets yields fewer than 200,000 available target samples due to severe sample imbalance\textemdash mainly ship detection datasets. 
As the inaugural phase of ATRNet, we focus on vehicle targets due to:
(1) Recent datasets predominantly are targets based on spaceborne SAR with constrained imaging geometries, whereas airborne platforms offer higher resolution (0.1-0.3m) and flexible imaging conditions in complex scenes.
(2) Vehicle recognition has driven SAR ATR research for three decades since MSTAR in the 1990s, establishing mature research systems.
(3) The 1990s-era MSTAR dataset, despite its seminal role, suffers from idealized acquisition conditions and saturated performance (near 99\% accuracy), failing to reflect real-world complexities and support innovation in the 2020s.

\begin{table}[!tb]
\centering
\footnotesize
\caption{\textbf{MSTAR v.s. ATRNet-STAR.} 
We can see the potential of the ATRNet-STAR dataset as a new benchmark, surpassing MSTAR in terms of data diversity. AFRL: Air Force Research Laboratory. NUDT: National University of Defense Technology. \# Types.: Number of object types. \# Samples: Number of samples.}
\label{table:vs}
\renewcommand\arraystretch{1.2}
\begin{tabular}{lcc} 
\toprule
\textbf{Dataset} & \textbf{MSTAR} & \textbf{ATRNet-STAR} \\ 
\cmidrule(lr){1-3}
Year & 1995 & 2025 \\
Country & USA & CHN \\
Institution & AFRL & NUDT \\
\# Types & 10 & 40 \\
Location & ideally centered & random, non-centered \\
Scene & grass & 5 realistic scenes \\
Platform & Airborne & Airborne \\
Resolution (m) & 0.3 & 0.12 $\sim$ 0.15 \\
Band & X & X, Ku \\
Polarization & single & quad \\
Depression angle ($^{\circ}$) & 15, 17, 30, 45 & 15, 30, 45, 60 \\
Azimuth angle ($^{\circ}$) & 0 $\sim$ 360 & 0 $\sim$ 360 \\
Annotation & classification & classification, detection \\
\# Samples & 14,557 & 194,324 \\
\bottomrule
\end{tabular}
\end{table}

\textbf{\textcolor{black}{Our Objective and Solution -}}
As a first step towards building a large-scale SAR target dataset, we introduce \textbf{ATRNet-STAR}, a fine-grained dataset for vehicle target recognition as a new breakthrough over the previous benchmark dataset MSTAR in Table~\ref{table:vs}. We completed program designation, data acquisition and processing, and benchmark construction with nearly two years of effort and considerable resources. In particular, ATRNet-STAR collects 194,324 object images from 40 target types, 5 scenes, and various imaging conditions with detailed annotation. It is the largest public SAR vehicle recognition dataset, even 10 times the size of any previous vehicle dataset, as shown in Table~\ref{table:atr_dataset}. 
Its sufficient target samples can support various aspects of generation, detection, and classification research.
To facilitate innovation and comparisons, we built a well-designed classification and detection benchmark, including 15 representative methods based on 7 experimental settings for robust ATR, few-shot, and transfer learning. 
Our results show that SAR ATR under complex conditions is still difficult, and representative methods that performed well in previous datasets cannot solve our difficult conditions, such as image angle and scene variations. We also find the effectiveness of the foundation model compared to others, which illustrates the importance of a large-scale dataset. Besides, the transfer learning results of various models pre-trained on this dataset to other datasets show that our dataset contributes to the feature representation of different ground targets. 
Finally, we provide detailed analyses and discussions to explore important issues for future research. We believe that our new dataset, with comprehensive imaging conditions and experiment benchmarks, will further promote the promise and reproducible research in SAR ATR.

\textbf{\textcolor{black}{Our Contributions -}}
The main contributions of this paper are listed below.

\textit{Dataset and Benchmarks -}
\textcolor{black}{To the best of our knowledge, the ATRNet-STAR is the largest dataset for SAR vehicle recognition. Building ATRNet-STAR is a challenging task and has taken nearly two years. We derive benchmark datasets from ATRNet-STAR to facilitate the research innovation and performance evaluation of target classification and detection methods.}




\textit{Code library -} 
\textcolor{black}{It is worth pointing out that we are the first to integrate publicly available and representative methods into a single code library to facilitate reproducible and large-scale performance evaluations in this field.}

\textit{Performance Evaluation -}
\textcolor{black}{We review most major approaches and provide a comparative study of 15 representative methods with 7 experimental settings on classification and detection benchmarks, named ATRBench, offering official baseline references when developing new approaches.}

\textit{Insights and Future Directions -}
\textcolor{black}{By analyzing the quantitative results, we provide insightful discussion that can guide
the development of SAR ATR algorithms and provide potential future research directions in this field.} 

\textcolor{black}{This work mainly focuses on the introduction, performance evaluation, and potential value of ATRNet-STAR. 
The annotated dataset, code library, and reported results are available at \url{https://github.com/waterdisappear/ATRNet-STAR}.}

The remainder of this paper is organized as follows. Sec.~\ref{sec:related work} discusses related work in SAR  target classification datasets and algorithms. Sec.~\ref{sec:dataset} introduces our dataset ATRNet-STAR. Secs.~\ref{sec:Benchmarks} conduct extensive experiments to build the benchmark. Sec.~\ref{sec:Conclusion} concludes the paper and discusses future work.



\section{Related Work}
\label{sec:related work}
Here, we review the datasets and algorithms for SAR target classification and finally discuss the significance of the datasets.

\subsection{Datasets for SAR target classification}
Datasets are the cornerstone of the development of new ATR techniques, especially in the era of deep learning. In recent years, there has been notable progress in RS datasets in various modalities~\cite{xiong2024earthnets,zhou2025fiftyyearssaratr,xu2024cloudseg}, including visible light~\cite{bastani2023satlaspretrain,li2025star,feng2023learning,ding2022object,zheng2023farset}, SAR~\cite{li2024sardet,wu2024fair}, infrared~\cite{RGBT-Tiny}, and multimodal~\cite{luo2024mmm,li2024sm3det}. However, not all modalities possess sufficient target samples to fully leverage the potential of big data and foundation models. In the past decades, the MSTAR~\cite{MSTAR} dataset has played a key role in SAR target classification tasks. However, current deep learning-based methods are saturated in this benchmark dataset setting. Insufficient samples in MSTAR also have data biases such as background correlation~\cite{kechagias2021automatic, li2023discovering, li2022research, 10820515}. Therefore, many datasets have been developed to promote the advancement of this field, especially in recent years. We analyze SAR target classification datasets that focus on individual target characteristics from the 1990s to the 2020s, which are detailed listed in Table~\ref{table:atr_dataset}. \textcolor{black}{Existing SAR target classification datasets face challenges such as no open source, insufficient samples, long-tailed distributions, and single collection conditions. }

\begin{table*}[!tb]
\centering
\footnotesize
\caption{\textbf{Statistics of the SAR target classification datasets from the 1990s to 2020s.} 
We focus on the electromagnetic scattering properties of individual objects, \emph{i.e.}, the most relevant to the classification dataset and task. OA.: Open access. \# Classes.: Number of object classes. \# Types.: Number of object types. Img. Size: image size. \# Img.: Number of images. Res.: Resolution. Pol.: Polarization. ``-'' refers to unknown. }
\label{table:atr_dataset}
\renewcommand\arraystretch{1.2}
\resizebox{0.95\linewidth}{!}{%
\begin{tabular}{cccccccccccl} 
\toprule
\makecell[c]{\textbf{Dataset}} & \textbf{Year} & \textbf{OA} & \textbf{Source} & \textbf{Band} & \textbf{Pol.} & \textbf{\# Classes} & \textbf{\# Types} & \textbf{Res. (m)} & \textbf{Img. Size} & \textbf{\# Img.} & \textbf{Description} \\ 
\cmidrule(lr){1-12}
MSTAR~\cite{MSTAR} & 1995 & $\checkmark$ & airborne & X & single & 8 & 10 & 0.3 & 128$\sim$193 & 14,557 & Most cited vehicle dataset \\
QinetiQ~\cite{QinetiQ} & 2004 & - & airborne & X & quad & - & 9 & 0.3 & 100$\sim$150 & 4,006 & Non-ideal background conditions \\
CV Domes~\cite{dungan2010civilian} & 2010 & $\checkmark$ & simulation & X & quad & 3 & 10 & 0.3 & - & - & 3D simulation civil vehicle \\
Gotcha~\cite{dungan2012wide} & 2012 & $\checkmark$ & airborne & X & - & 7 & 13 & 0.3 & - & - & 3D civil vehicle dataset \\
SARSIM~\cite{malmgren2017improving} & 2017 & $\checkmark$ & simulation & X & - & 7 & 14 & 0.1 & 139 & 21,168 & Simulation CAD vehicle \\
OpenSARShip~\cite{huang2018opensarship} & 2018 & $\checkmark$ & satellite & C & dual & 16 & - & 2.7$\sim$22 & 9$\sim$445 & 26,679 & Sentinel-1 ship dataset \\
HR4S~\cite{meng2019construction} & 2019 & - & satellite & C & quad & 21 & - & 3$\sim$25 & - & 1,962 & GF-3 and RadarSat-2 ship \\
MGTD~\cite{belloni2017sar} & 2019 & - & laboratory & X & single & 1 & 2 & 0.01 & 128 & 1,728 & Scaled vehicle models \\
SAMPLE~\cite{lewis2019sar} & 2019 & $\checkmark$ & simulation & X & single & 7 & 10 & 0.3 & 128 & 2,690 & Simulation and measured pairs \\
FUSAR-Ship~\cite{hou2020fusar} & 2020 & $\checkmark$ & satellite & C & dual & 98 & - & 1.1$\sim$1.7 & 512 & 5,243 & High-resolution GF-3 ship \\
IRIS-SAR~\cite{ahmadibeni2020aerial} & 2020 & - & simulation & - & - & 8 & 355 & - & 512 & 63,900 & Simulation CAD models \\
NUAAminiSAR~\cite{zhu2022SAR} & 2022 & - & airborne & X & - & 9 & 9 & 0.1 & - & - & Multi-angle circular SAR vehicle \\
MATD~\cite{wang2022multiangle} & 2022 & $\checkmark$ & airborne & Ku & - & 2 & 2 & - & 128 & 144 & Multi-angle stripmap SAR aircraft \\
SAR-ACD~\cite{sun2022scan} & 2022 & $\checkmark$ & satellite & C & single & 2 & 6 & 1 & 32$\sim$200 & 3,032 & GF-3 aircraft dataset \\
\cmidrule(lr){1-12}
\textbf{ATRNet-STAR} & 2025 & $\checkmark$ & airborne & X, Ku & quad & 21 & 40 & 0.12 $ \sim $ 0.15 & 128 & 194,324 & A large-scale fine-grained dataset\\
\bottomrule
\end{tabular}
}
\end{table*}


\textbf{Vehicle datasets -}
Vehicle datasets usually have various fine-grained vehicle types and diverse imaging conditions designed to evaluate robust classification performance under varying training and test sets. 
MSTAR was released by the Defense Advanced Research Projects Agency (DARPA) as the first public dataset for SAR target classification. It contains SAR image slices of 10 military vehicles and a reference target under different imaging angles with spotlight SAR. However, MSTAR has a background correlation problem and ideal collection conditions. 
Compared to MSTAR, the QinetiQ~\cite{QinetiQ} dataset provided a non-idealized acquisition condition with a non-centered target location and stronger background clutter. 
CV Domes~\cite{dungan2010civilian} and Gotcha~\cite{dungan2012wide} constructed simulation and measurement samples of civilian vehicles. These two datasets are in the raw echo data format and can be used for SAR imaging research, but they are not as convenient for researchers to use as MSTAR. Besides, these two civilian vehicle datasets are not rich in target classes and acquisition conditions. 
SARSIM~\cite{malmgren2017improving} and SAMPLE~\cite{lewis2019sar} are two simulation datasets designed to study transfer learning. In particular, SARSIM uses 14 CAD models, and SAMPLE has simulated and measured sample pairs based on partially MSTAR's vehicle. The simulation data have discrepancies with the measured data in background clutter, target scattering characteristics, and speckle noise. 
Although MSTAR includes 3 grasslands, most public images are from the same location. MGTD~\cite{belloni2017sar} collects training and test sets in two different laboratory backgrounds. 
Another issue is that the depression angles of the MSTAR public images are mostly in 15$^{\circ}$ and 17$^{\circ}$, with some classes missing in other depression angles. Hence, NUAAminiSAR~\cite{zhu2022SAR} collects images of military targets at different depression angles by circular SAR in a national defense park, including 8 military vehicles and an airplane. 
As shown above, existing vehicle datasets lack sufficient diversity in target types, scene settings, and sensor conditions.

\textbf{Ship datasets -}
SAR ship datasets have emerged with the development of satellite SAR technology. The difference from the vehicle dataset is that the ship classification dataset usually uses a satellite platform, and the imaging angles are limited. Ship datasets can be constructed and annotated using global ship tracking systems, but label noise remains a significant challenge.
OpenSARShip~\cite{huang2018opensarship}, HR4S~\cite{meng2019construction}, and FUSAR-Ship~\cite{hou2020fusar} advance the development of ship classification methods. These datasets are derived from different satellites and face difficult harbor backgrounds and class imbalance challenges. 
OpenSARShip collects medium-resolution images of 16 classes in harbors with Sentinel-1. 
HR4S builds high-resolution ship images in ports and coastal areas with GF-3 and RadarSat-2. 
FUSAR-Ship contains high-resolution images under various sea backgrounds with 15 main classes and 98 subclasses of ocean targets based on GF-3.
However, OpenSARShip and FUSAR-Ship have serious class imbalance and long-tailed distribution challenges. Researchers constructed various experimental settings and only used some target classes for classification tasks.

\textbf{Aircraft datasets -}
Aircraft datasets have emerged in recent years, presenting new challenges in SAR ATR. Due to their streamlined surfaces and less obvious scattering characteristics, aircraft are more difficult to classify than vehicle and ship targets.
Multiangle Aircraft Target Dataset (MATD)~\cite{wang2022multiangle} collects SAR images of two aircraft at different imaging angles with a drone stripmap SAR equipment. 
SAR-ACD~\cite{sun2022scan} contains 2 classes and 6 types of civil aircraft in 3 airports. 

\textbf{Others -}
IRIS-SAR~\cite{ahmadibeni2020aerial} is rich in aircraft, ships, and vehicle simulation types. It considers noise, clutter, and shadows in SAR image simulation. However, it does not take into account occlusion in complicated backgrounds.

\textbf{Object detection dataset -}
Beyond the classification datasets focused on individual targets, numerous SAR object detection datasets are designed to detect multiple targets. These datasets typically feature large image sizes, significant background interference, and complex targeting challenges. Examples include SARDet-100K~\cite{li2024sardet} and FAIR-CSAR~\cite{wu2024fair}, mainly satellite-based datasets containing approximately 100,000 images with resolutions exceeding 1 meter. This limited resolution and imaging angle of satellite makes it hard to detect and classify small targets, such as vehicles, in complex scenes. Non-cooperative objectives make it difficult to label the target types and accurately control the characteristics of the target under different angles and scenes. Therefore, our dataset provides a more fine-grained SAR vehicle characterization dataset on target class and imaging conditions.

Collecting target samples under different imaging conditions remains a challenging and resource-intensive endeavor. After the above investigation, we suggest that a good SAR ATR dataset should pay attention to the following key points.

\begin{enumerate}
\item[$\bullet$] 
\emph{Richness -} A dataset should include diverse target classes, scenes, and imaging conditions to reflect real-world environments and reduce biases. It also needs to contain enough formats and metadata, such as complex data and imaging parameters.
\item[$\bullet$] 
\emph{Standardization -} An official dataset and benchmark are beneficial to fair research and comparisons. In addition, the dataset format and target system should be standardized to be compatible with mainstream datasets and for further expansion and scaling.
\item[$\bullet$] 
\emph{Usability -} A dataset should be readily accessible and user-friendly for researchers to download and utilize. Some datasets may be difficult to access or require additional SAR imaging and pre-processing steps.  
\end{enumerate} 

\begin{figure*}[!tb]
\centering
\includegraphics[width=19.2cm]{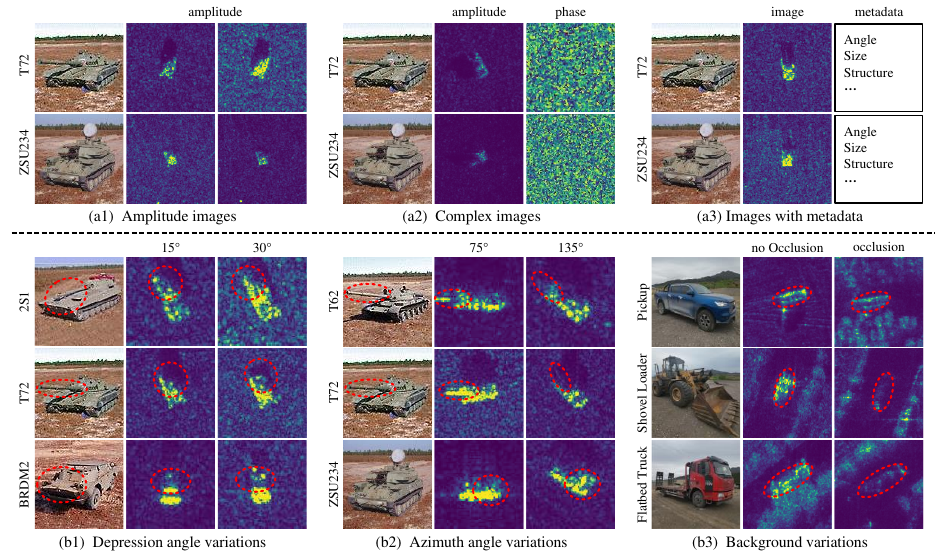}
\caption{\textbf{Importance of Dataset.} Taking the MSTAR dataset as an example, \textbf{subfigures (a1)-(a3)} describe the common data inputs of existing algorithms. The MSTAR dataset provides processed magnitude data, raw complex data, and corresponding metadata. \textbf{Subfigures (b1)-(b2)} show its mainstream experimental settings that are based on various imaging conditions of the MSTAR dataset. Other experimental settings include target configuration/version variation and simulation settings (simulation noise erosion and occlusion)~\cite{10283916}. However, the limited MSTAR acquisition conditions restrict the further study of the SAR ATR. For example, the current occlusion usually uses zero values to fill some image regions, but the real occlusion in our dataset \textbf{subfigure (b3)} is not the same as the existing simulation. (pseudo-color for better visualization of SAR magnitude images, and subfigures (b) are cropped and centered for better visualization of target signature variations.)}
\label{fig_prblem}
\end{figure*}

Therefore, we open-source a new large-scale dataset, ATRNet-STAR, with various classes and imaging conditions. This dataset is carefully labeled with extensive metadata and has an official benchmark based on 7 experimental settings and 15 comparison methods with 2 data formats. This dataset is our first step in building a new SAR target dataset system. In the future, we will expand the objects from vehicles to a wider category list to further promote SAR ATR research.

\subsection{Review of SAR Target Classification Algorithms}
Early target classification relies on hand-crafting features, while contemporary methodologies are predominantly anchored in deep learning-based methods. According to their different feature representations, we divided target recognition methods into four distinct paradigms: traditional features, deep learning, feature fusion, and classification with metadata. In an era increasingly dominated by data-driven algorithms, the availability of a rich, standardized, and usable dataset is becoming increasingly crucial.

\textbf{Traditional features -} 
Previous studies have proposed a wide range of valuable features, encompassing geometric structure features~\cite{ikeuchi1996invariant}, electromagnetic-scattering features~\cite{potter1997attributed, zhang2023scatterer, li2019sar, deng2024polsar}, time-frequency features~\cite{spigai2011time}, local descriptors~\cite{dong2020keypoint,song2016sar}, and sparse representations~\cite{hou2015sar}. These features are particularly tailored to the special coherent imaging mechanisms. For instance, SAR-HOG~\cite{song2016sar} employs a ratio-based gradient computation rather than the traditional differential gradient approach, and Attributed Scattering Centers (ASC) provides a feature set including location, geometry, and polarization information. Notably, extracting electromagnetic-scattering features relies on complex images with phase information, underscoring the necessity of providing comprehensive data formats.

\textbf{Deep learning -} 
Deep learning algorithms can uncover intricate patterns within datasets, particularly when ample data is available. Nowadays, these algorithms achieve remarkable accuracy, reaching nearly 99\% on MSTAR’s basic Standard Operating Condition (SOC) settings. However, the Extended Operating Condition (EOC), which includes robustness and few-shot, remains a significant challenge.
Convolutional Neural Network (CNN)~\cite{lecun1998gradient, yuan2025strip, liu2022convnet} is a common network in SAR target classification. 
All-Convolutional Network (A-ConvNet)~\cite{chen2016target} was proposed to reduce overfitting and had high accuracy under many experiment settings but is not robust to noise. In recent years, Transformer~\cite{vaswani2017attention,dosovitskiy2020image,liu2024vst} emerged as a powerful alternative, offering new perspectives on feature extraction. 
After a decade of significant progress, SAR target recognition now incorporates many advanced deep learning techniques such as data augmentation~\cite{lewis2019realistic, geng2023target}, attention mechanism~\cite{ren2021extended, wang2022sar}, domain alignment~\cite{10283916}, reinforcement Learning~\cite{jiang2024azimuth,10559836}, transfer learning~\cite{yang2023sar}, meta-learning~\cite{fu2022few,yang2024dynamic,he2025simulation}, physical deep learning~\cite{datcu2023explainable,10568211,huang2022physically}, semi-supervised learning~\cite{zhang2023semi}, and self-supervised learning~\cite{zhai2022weakly,li2024predicting}. The majority of research endeavors are based on the MSTAR dataset, and a subset of studies also create the different non-open source settings of OpenSARShip and FUSAR-Ship datasets. Nonetheless, the reliance on outdated datasets and the lack of standardized benchmarks are impeding the advancement of SAR target recognition in the 2020s.

\textbf{Feature fusion -} 
In order to improve the robustness and interpretability of deep learning methods, many feature fusion methods have been proposed combining traditional and deep features. 
Kechagias~\emph{et al.}~\cite{kechagias2018fusing} discussed the fusion of deep learning and sparse coding.
FEC~\cite{zhang2020fec} used deep CNN combined with ASC features derived from complex data, incorporating both amplitude and phase information. 
Li~\emph{et al.}~\cite{li2022novel} combined scattering centers (SC) with graph convolutional network. 
Besides feature fusion methods, ESF~\cite{feng2022electromagnetic} applied ASC features to guide deep model training. 
EFTL~\cite{liu2022eftl} proposed using the ASC model as weight initialization of a complex convolutional network. PIHA~\cite{huang2024physics} proposed a hybrid attention mechanism with ASCs and the deep learning framework. 
LDSF~\cite{10753051} constructed a heterogeneous graph for the first time to fully exploit both the scattering information of the target components and their interactions.
However, many SAR target datasets only provide magnitude images, ignoring phase information and quantization issues.

\textbf{Classification with metadata -}
Imaging angles and target-related metadata are the two most prevalent categories of ancillary data. MSTAR and SAMPLE datasets furnish the imaging angles for each respective image.
RotANet~\cite{wen2021rotation} combined predicting the rotational pattern of an image sequence with self-supervised learning. 
Zhang~\emph{et al.}~\cite{zhang2024optimal} investigated the problem of optimal image imaging angle selection in few-shot recognition based on the MSTAR dataset.
AaDRL~\cite{jiang2024azimuth} explored the SAR active recognition framework to fully exploit target characteristic variations at different azimuth angles based on the SAMPLE dataset. 
In addition, multi-view methods~\cite{tong2023active} were proposed with image sequences from different imaging angles. 
Since the target types are known, the researchers used 3D models and shape parameters as metadata for fine-grained tasks. 
For example, SARBake~\cite{malmgren2015convolutional} used 3D CAD models and projection relations for semantic segmentation, and DBAE~\cite{wei2023zero} utilized size and structure attributes for zero-shot classification.
Consequently, a SAR target dataset enriched with pertinent metadata, encompassing both imaging parameters and detailed target characteristics, becomes an indispensable step in pursuing accurate and robust SAR target recognition.

\textbf{Experimental settings -} 
Indeed, the MSTAR dataset has more than 50 types of experimental configurations for SOC and EOC, as detailed in~\cite{kechagias2021automatic}. These configurations include image angle variation, target version/configure variation, noise corruption, occlusion, resolution variation, and background clutter variation~\cite{kechagias2021automatic,10283916}. In a similar vein, other SAR classification datasets, such as OpenSARShip and FUSAR-Ship, lack a standard experimental benchmark, and researchers also devise different bespoke settings that leverage subsets of the target class for classification~\cite{huang2022physically}.

\textbf{Evaluation metrics -} 
Classification accuracy and confusion matrix are the primary metrics employed for evaluating the performance of classification models. Accuracy quantifies the proportion of correctly classified samples relative to the total number of samples in the dataset. The confusion matrix provides a detailed correspondence between the predicted and true class labels, with each column representing the predicted class and each row representing the true class. Besides, the Receiver Operating Characteristic (ROC) curve is a valuable tool for assessing used for binary classification and detection tasks such as suppressing background clutter and outlier rejection~\cite{chen2016target}.

We noticed a pronounced preference for the MSTAR dataset among SAR classification algorithms. Most algorithms are based on the MSTAR dataset, with others using datasets such as SAMPLE, OpenSARShip, and FUSAR-Ship. This preference is attributed to the distinctive benefits of the MSTAR dataset released in the 1990s. As depicted in Fig.~\ref{fig_prblem} (a), studies requiring complex data have opted for the MSTAR dataset, whereas other datasets that only provide magnitude data cannot be used to validate their algorithms. In addition, information about vehicle types in the dataset is also available on the web. The MSTAR dataset has a more balanced target sample and diverse imaging conditions compared to other datasets with long-tailed distributions. Therefore, MSTAR has accumulated a large number of algorithms and experimental settings. However, this dataset has not released all target samples, and many algorithms have achieved peak performance in common popular settings~\cite{kechagias2021automatic}, such as classification with ten target types, imaging angle variation, and version variation. The ideal acquisition conditions of the MSTAR dataset cannot show that the recognition problem is truly well solved. Scholars aim to address these limitations by proposing novel simulation experimental settings. However, as shown in Fig.~\ref{fig_prblem} (b), some simulation experimental settings may not accurately reflect the real-world conditions. Besides, ensuring sample diversity is challenging due to the high data collection and labeling costs. These limitations collectively impose significant constraints on the advancement of SAR ATR techniques. Constructing a new open-source dataset benchmark with diverse target classes and imaging conditions is an urgent cornerstone of excellent and stable SAR ATR techniques.

\section{\textcolor{black}{Building ATRNet-STAR: Why, What, How}}
\label{sec:dataset}

In this section, we present and discuss ATRNet-STAR from three perspectives: (1) \textbf{Why} we constructed this dataset in the 2020s; (2) \textbf{What} we did to ensure its diversity and standardization; (3) \textbf{How} the diversity and statistical analysis of this dataset to support SAR ATR.

\begin{figure*}[!tb]
\centering
\includegraphics[width=0.95\textwidth]{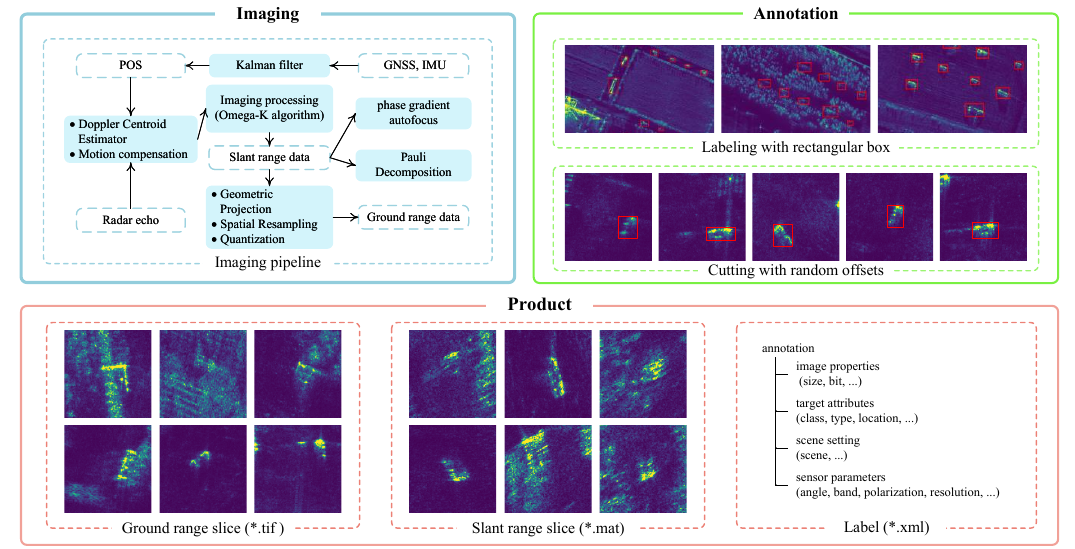}
\caption{\textbf{Illustrations of data acquisition.} 
We annotate and cut to build target slices with corresponding metadata information. The range dimension of slant range complex data is in the line of sight direction, which results in the deformation of the target shape in this dimension. Therefore, we also provide ground range images \textcolor{black}{and} slant range amplitude data.}
\label{fig_acquisition}
\end{figure*}

\begin{table*}[!tb]
\centering
\footnotesize
\caption{\textbf{Acquisition parameters of open-source ATRNet-STAR.} 
We segment the targets with fixed-size slices and random offsets to investigate their characteristics and aim for robust recognition under different collection conditions. We varied the target location in urban and factory scenes and collected data at the factory several times. In addition, some targets cannot be labeled due to heavy occlusion.
\textcolor{black}{The carrier frequency of the X-band is 9.6 GHz, and that of the Ku-band is 14.6 GHz, both with a bandwidth of 1200 MHz. The noise figure for both bands is typically under 4 dB. The antenna gains are 12 dB for the Ku-band and 14 dB for the X-band. The platform flight speed is 10 m/s, and the pulse repetition frequency is 124.998 Hz. Images at different azimuth angles are acquired by flying different regular polygons around the scene center. The flight altitude corresponds to different depression angles. Most SAR data acquisitions were conducted during nighttime hours (20:00–02:00 Beijing time), under wind conditions averaging around Force 2 on the Beaufort scale and in the absence of precipitation.}
\# Types.: Number of object types. Res.: Resolution. Pol.: Polarization. Dep.: Depression angle. Azi.: Target azimuth angle interval. Img. Size: image size. \# Img.: Number of images in ground and slant range coordinate systems.}
\label{table:statistics of our dataset}
\renewcommand\arraystretch{1.2}
\begin{tabular}{ccccccccccccc} 
\toprule
\textbf{Scene} & \textbf{\# Types} & \textbf{Platform} & \textbf{Mode} & \textbf{Band} & \textbf{Res. (m)} & \textbf{Pol.} & \textbf{Dep. ($^{\circ}$)} & \textcolor{black}{\textbf{Altitude (m)}} & \textbf{Azi. ($^{\circ}$)} & \textbf{Img. Size} & \textbf{\# Img.} \\ 
\cmidrule(lr){1-12}
City & 40 & airborne & stripmap & X & 0.12 $ \sim $ 0.15 & quad & 15, 30, 45, 60 & 150, 300, 300, 400 & 5 & 128 & 83,465 \\
Factory & 40 & airborne & stripmap & X \& Ku & 0.12 $ \sim $ 0.15 & quad & 15, 30, 45, 60 & 120, 300, 320, 400& 30 & 128 & 63,597 \\
Sandstone & 40 & airborne & stripmap & X \& Ku & 0.12 $ \sim $ 0.15 & quad & 15, 30, 45, 60 & 120, 300, 300, 300& 30 & 128 & 30,720 \\
Woodland & 11 & airborne & stripmap & X \& Ku & 0.12 $ \sim $ 0.15 & quad & 15, 30, 45, 60 & 120, 300, 300, 300& 30 & 128 & 8,094 \\
Bare soil & 11 & airborne & stripmap & X \& Ku & 0.12 $ \sim $ 0.15 & quad & 15, 30, 45, 60 & 120, 300, 300, 300 & 30 & 128 & 8,448 \\
\bottomrule
\end{tabular}
\end{table*}

The emergence of new datasets and papers has led to rapid progress in various areas of SAR ATR, demonstrating the attention given to this issue in recent years. However, a large-scale dataset with various target classes and imaging conditions is still lacking because of SAR's expensive costs and annotation. We aim to achieve as much diversity as possible in target classes, scene settings, and sensor conditions. 
Such multidimensional design establishes it as an instrumental resource for advancing SAR ATR research. Its principal values for the SAR ATR are fourfold:
\begin{enumerate}
\item[$\bullet$] 
To provide a challenging and rich dataset for SAR vehicle recognition;
\item[$\bullet$] 
To systematically study the effects of different conditions on SAR target characteristics;
\item[$\bullet$] 
To give a benchmark to facilitate method innovation and comprehensive comparisons;
\item[$\bullet$] 
To promote the development of new ATR technologies and research topics.
\end{enumerate} 


\subsection{Construct program}
We aim to extend the dataset diversity as much as possible regarding targets, scenes, and sensors. These aspects are derived from the MSTAR's discussion~\cite{Ross1999SAR} on SAR target characteristics.

\textbf{Target -} 
First of all, according to the Chinese and European vehicle classification standards~\cite{GA802－2019, Passengercarclassification}, we confirm the main target \emph{classes} based on uses and structures of vehicles, including people carriers (car and bus), goods carriers (truck), and special purpose vehicles (special). Then, we subdivide each class into various \emph{subclasses} based on size and special structure and collect as many different \emph{types} as possible. Finally, ATRNet-STAR has 4 classes, 21 subclasses, and 40 types. It covers most civilian vehicle categories and has the most richness on civil target classes compared to other SAR vehicle datasets.

\begin{figure*}[!tb]
\centering
\includegraphics[width=0.92\textwidth]{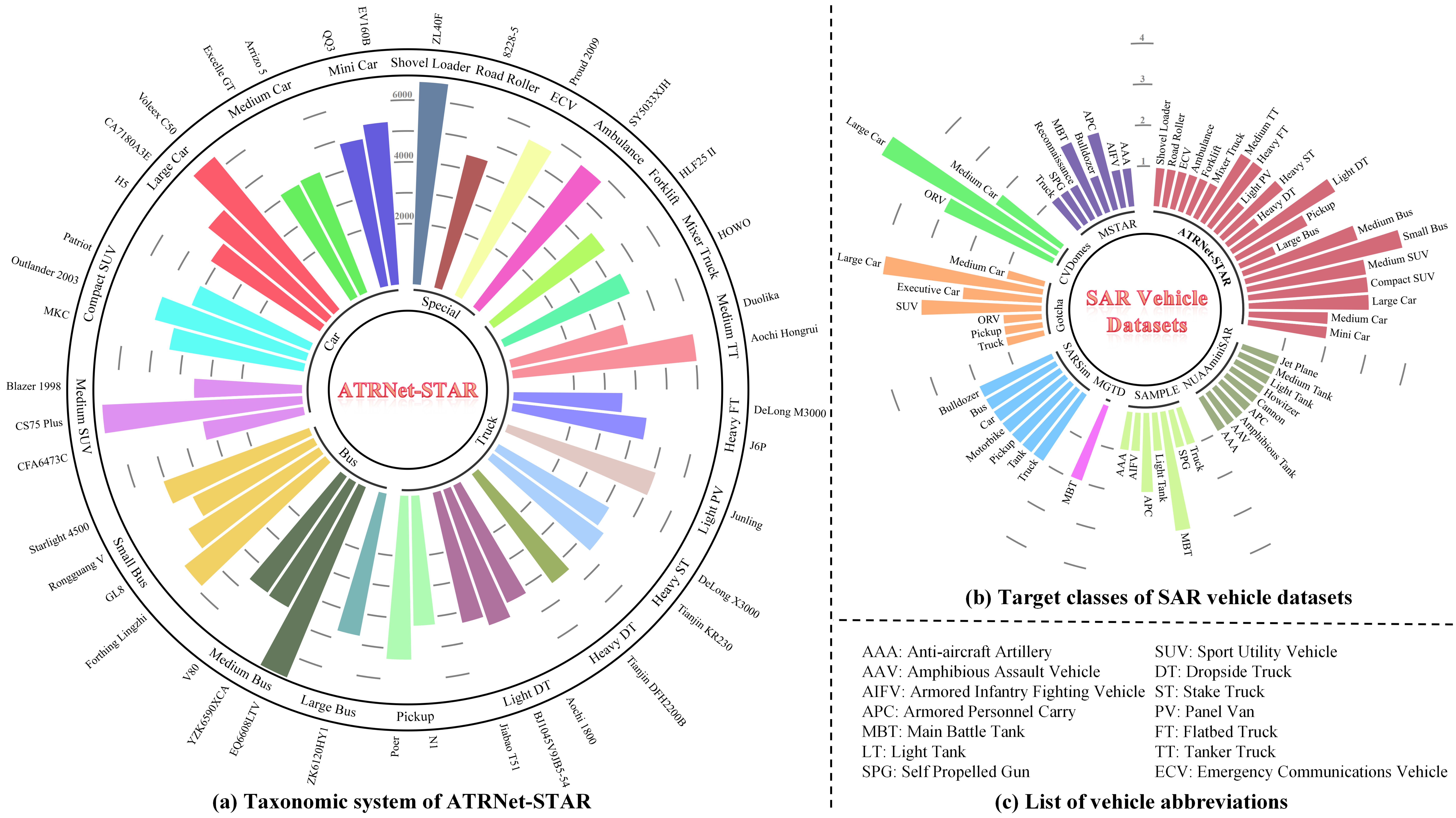}
\caption{
\textbf{Taxonomic systems.} 
For civilian vehicles, the taxonomy is based on Chinese and European vehicle classification standards~\cite{GA802－2019, Passengercarclassification}, according to the vehicle's purpose, structure, size, and mass. For military vehicles, we followed the MSTAR taxonomic system~\cite{Ross1999SAR}. 
\textbf{(a) Taxonomic systems of ATRNet-STAR.} 
Our dataset is a comprehensive civilian vehicle map covering 4 classes, 21 subclasses, and 40 types. We provide a detailed illustration of the histogram distribution for these 40 vehicle types.
\textbf{(b) Target classes of SAR vehicle datasets.} 
We statistically analyze the number of civilian and military vehicle classes and types in SAR vehicle datasets. 
\textbf{(c) List of vehicle abbreviations.}}
\label{fig_target_compare}
\end{figure*}

\begin{figure*}[!tb]
\centering
\includegraphics[width=0.90\textwidth]{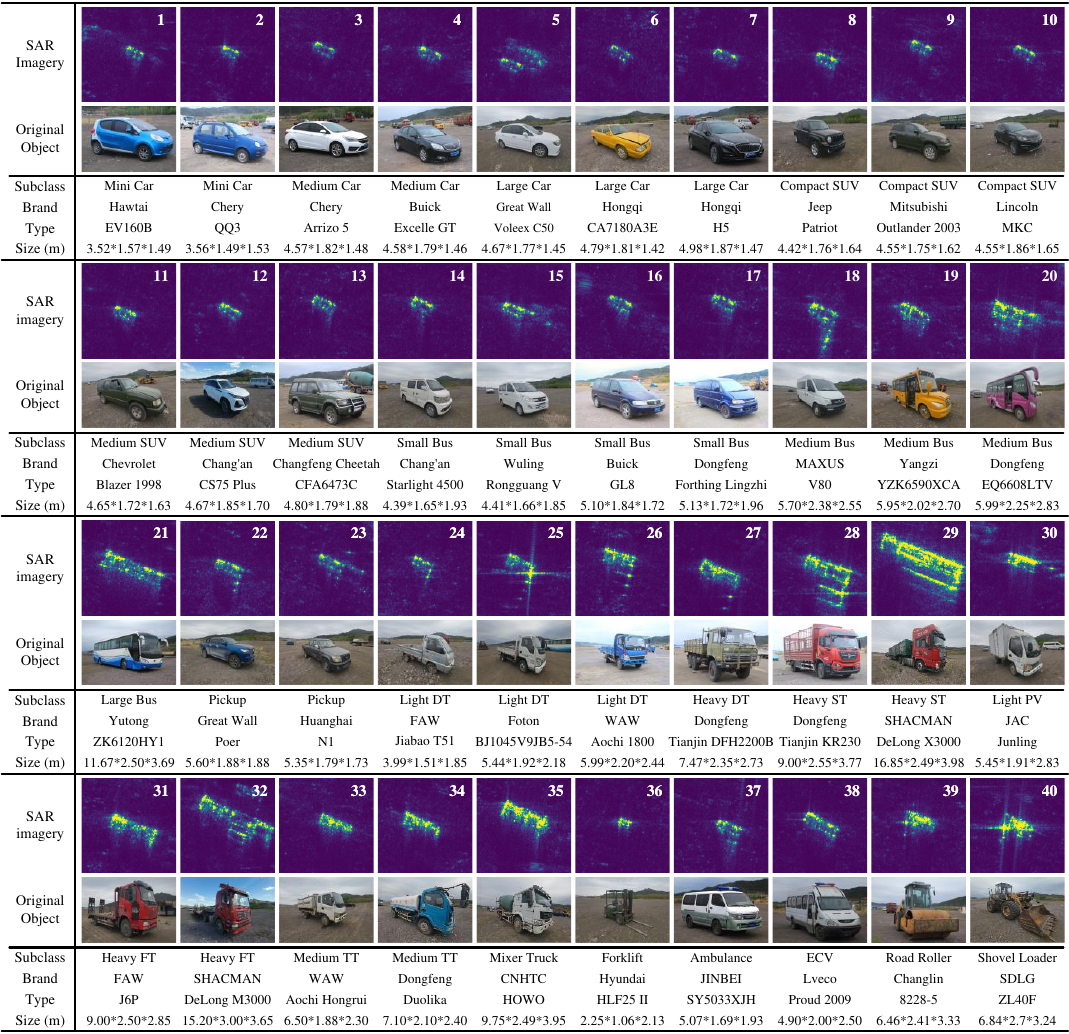}
\caption{\textbf{Demo images, subclasses, types, and sizes of 40 vehicles in ATRNet-STAR.} 
Here are the SAR and corresponding objects. Their size (length * width * height) measurements are listed in meters (m). As we can see, the obvious difference between the vehicles in the SAR images is the scattering characteristic variation due to size and structure. Therefore, we classify various vehicle types based on their size and structure. Besides, it is worth noting that our SAR images are not in the ideal situation where the target is located right in the center of the demo image, just as in the MSTAR dataset, but has a random offset similar to the QinetiQ dataset.}
\label{fig_target_category}
\end{figure*}


\textbf{Scene -} 
Most previous target datasets were collected in open scenes such as grasslands, focusing on target characteristics and lacking significant background interference. \textcolor{black}{This dataset is designed to collect data from both simple and complex scenes\footnote{Since we mainly consider single target characteristics, discussing densely distributed targets such as car parks are not included in this version.}, encompassing urban, factory, sandstone, woodland, and bare soil. Complex scenes (urban, factory, and woodland) contain interferences such as buildings and trees, whereas simple scenarios (sand and bare soil) are a cleaner background with grass or stone. This various environmental sampling strategy ensures that the acquired data more accurately reflects real-world recognition challenges in the wild.}


\textbf{Sensor -}
\textcolor{black}{The primary variations typically arise from imaging angles. Accordingly, this dataset incorporates a balanced azimuth sampling scheme across different depression angles. The azimuth sampling density in most scenes is less intensive than MSTAR, constrained by the operational costs of stripmap imaging, but maintains balanced sampling across all target types under varying depression angles. Furthermore, quad-polarization and multi-band data were acquired to investigate their impacts on SAR target characteristics. Simultaneously, some satellite-platform imagery was considered in coordination to examine platform-specific and resolution-related effects.}


Table~\ref{table:statistics of our dataset} presents the open-source list \textcolor{black}{and acquisition parameters}. After a year and a half's effort (from April 2023 to October 2024), we completed program designation, data processing, and benchmark construction. We commissioned the Institute of Stealthy Engineering Technology Company in Beijing to complete the data collection and annotation.

\subsection{Data acquisition}
Here, we present our data acquisition pipeline in terms of imaging, annotation, and product.

\textbf{Imaging -} 
We use the Unmanned Aerial Vehicle (UAV) platform to carry sensor equipment. Two antennas acquire quad polarization radar echo in the X and Ku bands. 
\textcolor{black}{As shown in Fig.~\ref{fig_acquisition}, POS data are derived by fusing GNSS and IMU measurements using a Kalman filter, supporting Doppler centroid estimation and motion compensation for echo data. The Omega-K algorithm is then applied to focus the time-domain signals into two-dimensional complex slant range images. These images undergo geocoding (geometric correction and projection to ground-range coordinates), followed by spatial resampling  and non-linear quantization to produce ground-range images. Additionally, phase gradient autofocus and Pauli decomposition are applied to the slant range images to generate other products.}


\textbf{Annotation -} 
Target classes and coordinates are annotated using rectangular box labels based on optical reference images and deployment records. Because of focusing on individual target signatures, all objects maintain specified separation distances during placement. Besides our vehicles, other vehicles in the scene are labeled as ``other''. After labeling, we acquired the target slices and added a random offset. 

\textbf{Product -} 
We offer data products in two coordinate systems. The distance dimension of the slant range data is the line of sight direction, and we provide original complex data. The ground range images is projected to the ground truth distance and is processed with nonlinear quantization. The corresponding annotation files include basic image information and target, scene, and sensor parameters.

\subsection{Professional system} 
\textcolor{black}{Here, we introduce professional systems to standardize the dataset and facilitate utilization.} We construct a target taxonomic system to facilitate subsequent extensions and integration with other datasets. Data format and annotation metadata are also discussed, considering the specificity of SAR imaging.

\textbf{Taxonomic system -} 
A taxonomic system is crucial when creating a large-scale fine-grained dataset~\cite{deng2009imagenet, hou2020fusar, Ross1999SAR}. Motivated by MSTAR's military taxonomic system~\cite{Ross1999SAR} and FUSAR-Ship's ship taxonomic system~\cite{hou2020fusar}, our hierarchical taxonomic system (\emph{class} $\rightarrowtail$ \emph{subclass} $\rightarrowtail$ \emph{type}) are based on vehicle's purpose, structure, size and mass referred to Chinese and European vehicle classification standards~\cite{GA802－2019, Passengercarclassification}. As shown in Fig.~\ref{fig_target_compare} (a), we first create 4 \emph{classes} based on the vehicle's different purposes, such as carrying passengers (car and bus), cargo (truck) and special purposes (special), and dual-purpose pickup is classified into the truck. Then, the system is further expanded into 21 \emph{subclasses} according to structural, size, and mass specifications. For example, cars are based on size and passenger capacity, and trucks depend on length, mass, and special structure. Finally, we collect the 40 vehicles \emph{types} (\eg EV160B and QQ3 types from different producers and brands in the mini car class.) with size information in Fig.~\ref{fig_target_category}. This hierarchical architecture enhances standardization in data acquisition protocols and supports modular dataset expansion.

\textbf{Data format -}  
Due to the unique imaging mechanisms, processing steps, and image properties of SAR, we offer different levels of products to support the comprehensive study of the impact of imaging steps on recognition problems. We provide magnitude images in the ground range coordinate system with 8-bit and 32-bit uint formats. The magnitude images with 16-bit uint and complex data with 32-bit float in the slant range coordinate system are also available. In subsequent experimental benchmarks, we use 8-bit ground distance magnitude images and 32-bit slant distance complex data. We also perform other processing based on the slant distance data, such as phase gradient autofocus and \textcolor{black}{Pauli} polarization decomposition, but the current version of this dataset does not include the processed data after these steps. Raw echoes are also stored, but due to the large storage ($\sim$25TB), they will not be available for upload and download. Researchers can contact us to obtain raw echo if needed. We hope these products will help the researchers easily work with our dataset while providing more tractability to explore and discuss specific imaging mechanisms and image properties for SAR recognition applications.

\textbf{Metadata system -} 
Considering the effects of target, scene, and sensor on SAR images, we add many factors to the annotation to facilitate various recognition methods with metadata. Image information includes basic image sizes and formats. The target attributes include taxonomic systems, target size, and image position. Although we discuss single target characteristics with classification tasks, rectangular boxes also support detection tasks. Sensor parameters have the depression angle, target azimuth angle, band, polarization, and resolution in two dimensions. However, we hope that researchers consider the difficulty and accuracy of obtaining metadata when making full use of them. For example, there is the potential for angular confusion when measuring target azimuths and the potential for vehicle size to change after modifications.

\subsection{Statistical Analysis}

We present key characteristics and comparative statistics of our dataset relative to others.

\textbf{Class distribution -} 
ATRNet-STAR dataset includes 4 classes, 21 subclasses and 40 types of vehicles with balanced samples, as shown in Fig.~\ref{fig_target_compare}. 
\textcolor{black}{Our dataset greatly enhances the civilian vehicle richness for the SAR ATR, compared to the measured dataset Gotcha and the simulated datasets CVDomes and SARSim. Its sufficient and balanced samples make it well-equipped to meet various experimental settings and studies.  
The length, width, and height of our 40 objects are listed in Fig.~\ref{fig_target_category}, and we can see that they have various structures, different sizes, and similar size ratios. }
Relative to other SAR target datasets, our dataset's enhanced diversity introduces novel formidable challenges to the SAR target fine-grained recognition research. Moreover, adequate samples guarantee the availability of a substantial array of target classes across experimental settings, facilitating more diverse investigations.


\textbf{Reference target -}  
Besides deploying corner reflectors for resolution measurements, we produce and place a reference target of multiple geometries as the ``SLICY'' target in the MSTAR dataset for scattering research.


\begin{figure}[!tb]
\centering
\includegraphics[width=8.4cm]{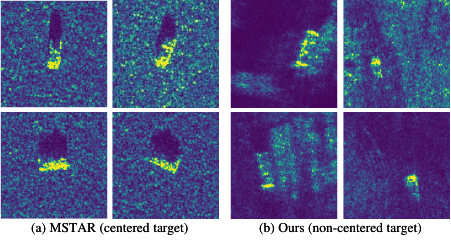}
\caption{\textbf{Non-centered target slices.} 
We apply the non-centered target setting following the QinetiQ dataset to increase detection difficulty.
}
\label{fig_center}
\end{figure}

\textbf{Non-centered location -} 
Most SAR target datasets place the target in the image center or contain only the target region. Since remote sensing differs from the human eye view that customarily centers the object of interest, the overhead view requires searching targets with more interference. We randomly added position offsets to increase the recognition difficulty with detecting non-centered target locations in Fig.~\ref{fig_center}. 

\begin{figure}[!tb]
\centering
\includegraphics[width=8.4cm]{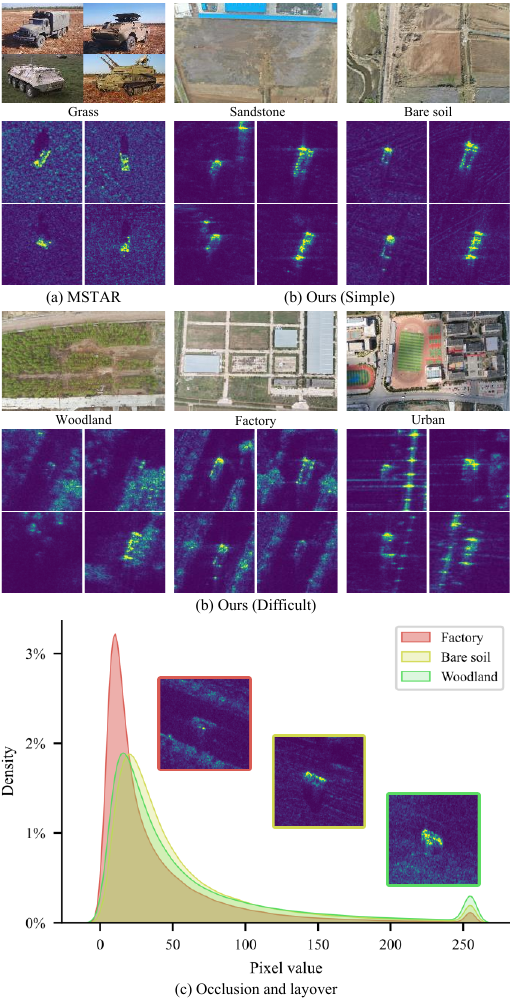}
\caption{\textbf{Influence of different scenes on target characteristics in SAR images.}
Most current SAR target classification datasets are collected in simple scenes. \textbf{(a) MSTAR} is collected in a flat grass scene with a clean background and pseudo-correlation between target and background. While \textbf{(b) our ATRNet-STAR} samples are collected under different positions in various scenes. In Fig. (b), it is clear that the same target under the same imaging angle has obvious signature changes due to different scenes. For example, other objects higher up in front of and behind the target can create occlusions and layovers, reducing reflected energy and increasing non-target scattering. Besides, the target shadow in difficult scenes is not as obvious as the MSTAR data. \textbf{(c) Occlusion and layover.}
Compared to the bare soil scene without interfering objects, the factory and woodland scenarios indicate occlusion and layover. Shadows from interfering objects in the former may obscure targets, while trees in the latter are likely to increase non-target scattering. We illustrate this problem with a single target demo at the same angle across scenes, but occlusion and layover result from a combination of target, interference, and imaging geometry. These statistics are from an SUV Chang'an CSCS75 Plus (Vehicle 12) at different image angles in the 3 scenes's ground range images.}
\label{fig_scene}
\end{figure}

\textbf{Occlusion in scenes -} 
Whereas previous vehicle datasets collect target samples in ideal environments, we systematically acquire samples across diverse scenes. Fig.~\ref{fig_scene} shows their different influences with occlusions and layovers in different scene regions. The shadow of buildings and roadside trees in the factory may obscure targets and reduce target scattering, and the nearby trees in woodland have an obvious \textcolor{black}{layover} and increase non-target scattering.

\begin{figure}[!tb]
\centering
\includegraphics[width=8.4cm]{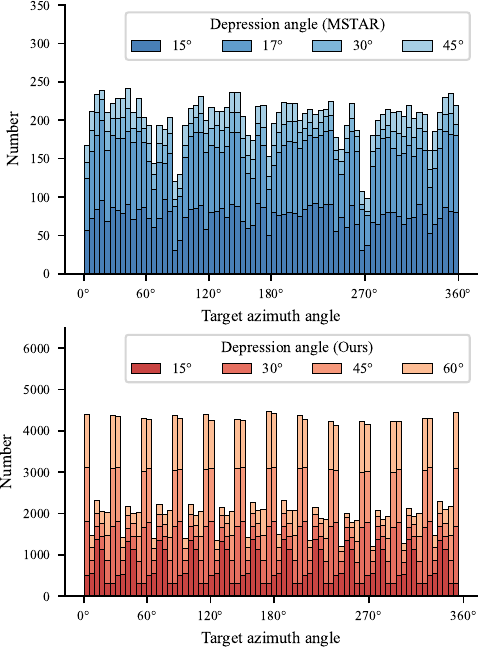}
\caption{
\textbf{Target azimuth angle distribution of (a) MSTAR and (b) ATRNet-STAR.} 
Compared to the MSTAR dataset, which has depression angles mainly publicized at 17$^{\circ}$ and 15$^{\circ}$ with incomplete angles for most target classes, our ATRNet-STAR dataset provides balance and comprehensive angles for all targets. However, the sampling interval of target angles is sparser due to stripmap imaging. 
}
\label{fig_target azimuth}
\end{figure}

\textbf{Imaging angle -} 
Different imaging angles affect the target scattering characteristics, and the imaging geometry relationship can change the interference in the scene. Our dataset has various imaging angles with balance distribution in different depression angles across scenes, as shown in Fig.~\ref{fig_target azimuth}. However, azimuth angle sampling density exhibits scene-dependent due to stripmap mode constraints and operational cost limitations.

\begin{figure}[!tb]
\centering
\includegraphics[width=8.4cm]{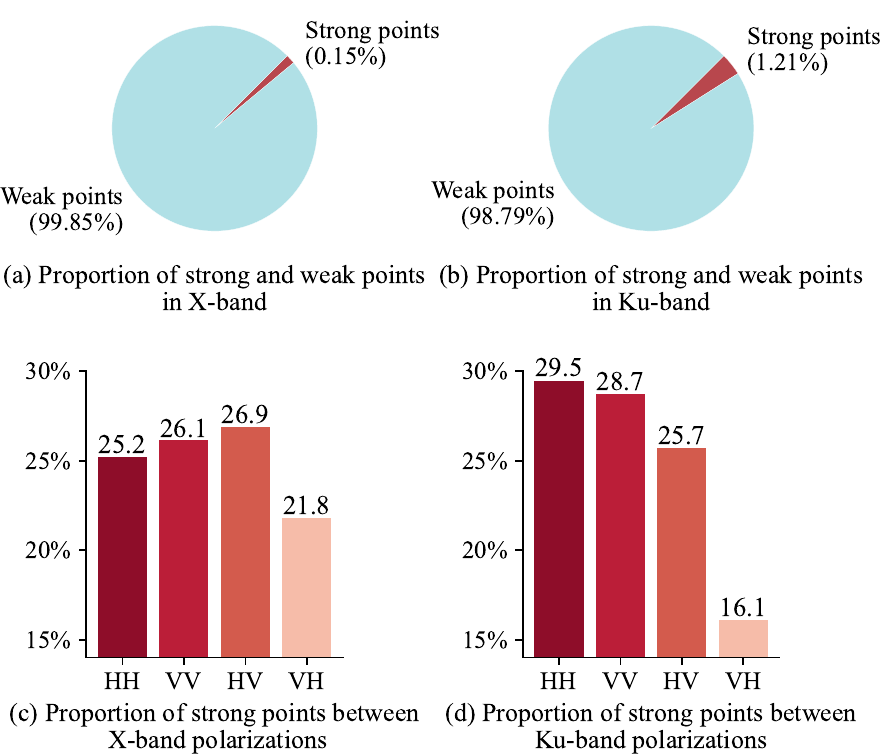}
\caption{
\textbf{Ratio of strong and weak points in different bands and polarization.} 
It can be noticed that the band and polarization can have a statistically significant effect on the target signatures. These statistical data are the target region inside the vehicle's rectangular box of sandstone scenes. Points with pixel values greater than or equal to 128 are treated as strong scattering points for 8-bit ground range images.
}
\label{fig_band}
\end{figure}

\textbf{Band and polarization -} 
Table~\ref{table:atr_dataset} shows most vehicle datasets with the MSTAR dataset's X-band and HH polarization settings. However, Fig.~\ref{fig_band} demonstrates that this setting is not the only one to consider, as band and polarization have a noticeable effect on target scattering. We provide two bands and quad polarization to support discussing these sensor parameters.

\textbf{Complex data -} 
Feature extraction based on complex data is a hot research concern, so we provide complex data in slant distance coordinates stored in \emph{.mat} format.


\textbf{Number of image bits -} 
RGB images are stored in 8-bit format, whereas SAR images have a larger range of digital data values. Therefore, we provide different bit formats to investigate quantification for enhancing weaker target scattering points.

\textbf{Cross-platform problem -} 
We collect a small amount of satellite data synchronously under the same scenes and targets to investigate the difficulty across platforms for future versions.

\textbf{Quality control -} 
To ensure high-quality data, we considered challenges such as target and site hire, airspace applications, weather conditions, target placing, and data checking. Eight people completed the data acquisition over six months. In addition, 14 labelers and two inspectors performed the data annotation for about four months. 
The entire project cost nearly two years.

\subsection{Dataset value}
Based on the diversity of target classes, scenes, imaging conditions, and data format with detailed annotation, we recommend that this dataset be used for SAR studies in the following areas.


\textbf{Robustness recognition -} 
Our dataset encompasses diverse acquisition conditions, with each image accompanied by exhaustive acquisition metadata. Therefore, we can discuss in detail the robustness of the SAR recognition algorithms when the acquisition conditions (\emph{\eg}, scene and sensor sensors) vary between the training and test sets.
If we consider data from different acquisition conditions as domains, the training set can include multiple domains to learn domain invariant features, \emph{i.e.}, the domain generalization~\cite{zhou2022domain}. 

\textbf{Few-shot learning -} 
Besides acquisition condition variation, the inherent challenge of limited training samples due to high SAR collection costs also presents critical research opportunities. The few-shot setting can be combined with robust recognition, and our dataset's large number of target types can enrich the task construction of meta-learning~\cite{hospedales2021meta,10264116,10533864,9760074} settings in existing few-shot learning.

\textbf{Transfer learning -} 
Our dataset contains a large diversity of samples that can be used to study the transfer problem of the pre-training model with different SAR target datasets. In addition, the large number of samples can further increase the volume of SAR target samples to advance the study of self-supervised learning~\cite{liu2021self,li2024predicting} and foundation models~\cite{bommasani2021opportunities,li2024saratrx}.

\textbf{Incremental learning -} 
The diversity of our dataset supports investigations into domain incremental learning and class incremental learning~\cite{van2022three}. The domain incremental learning can improve the algorithm's robustness with a dynamic process, and class incremental learning can progressively increase the ability to recognize or reject new classes. These incremental capacities are essential for SAR ATR in open environments.

\textbf{Physical deep learning -} 
SAR images have unique properties, such as complex phase and polarization. Our dataset provides multi-format data to facilitate recognition studies leveraging these intrinsic attributes instead of only relying on quantized SAR magnitude images. In addition, detailed metadata can also provide more gains for SAR ATR.

\textbf{Generative Models -} 
Beyond recognition tasks, our dataset enables controllable generation of target samples under diverse imaging conditions~\cite{huang2024generative}, as well as estimation of target parameters across varying acquisition scenarios.

We encourage researchers to propose new experimental settings and research issues based on this dataset. \emph{Don't hesitate to contact us if you get new ideas.}

\section{Benchmarks}
\label{sec:Benchmarks}

We consider 7 experimental settings with 2 data formats as classification and detection benchmarks from this dataset in Table~\ref{table:SOC and EOC settings}, named ATRBench. \textcolor{black}{And we select 15 representative classification and detection methods.} The data formats consist of magnitude images in the ground range coordinate system and complex images in the slant range coordinate system.

\subsection{Experimental settings}
Here, we describe the experimental settings in Table~\ref{table:SOC and EOC settings}, including 2 SOC settings sampled from similar distributions and 5 EOC settings with obvious condition and distribution shifts.

\begin{table*}[!tb]
\centering
\footnotesize
\caption{
\textbf{SOC and EOC settings of ATRBench derived from ATRNet-STAR. }
The imaging conditions in SOC are similar, while EOC considers variations in a single imaging condition. Simple scenes are sandstone and bare soil, and complex scenes are urban, factory, and woodland. The separate labeling of the ground distance and slant distance images results in their numbers of annotations not being strictly corresponding.
\# Types.: Number of object types. 
Dep.: Depression angle. 
Azi.: Target azimuth angle.
Pol.: Polarization.
\# Img. (Ground): Number of images in ground range.
\# Img. (Slant): Number of images in slant range. 
}
\label{table:SOC and EOC settings}
\renewcommand\arraystretch{1.1}
\begin{tabular}{lcccccccccc} \toprule
Setting  & Set & \# Types & Scene & Dep. & Azi. & Band & Pol. &  \makecell{\# Img.\\(Ground)} & \makecell{\# Img.\\(Slant)} \\ \cmidrule(lr){1-10}
\multirow{2}{*}{SOC-40} & train & \textbf{40} & all & 15, 30, 45, 60 & 0$\sim$360 & X \& Ku & quad & 68,091 & 67,780 \\
 & Test & \textbf{40} & all & 15, 30, 45, 60 & 0$\sim$360 & X \& Ku & quad & 29,284 & 29,169 \\ \cmidrule(lr){1-10}
\multirow{2}{*}{SOC-50} & train & \textbf{50} & all & 15, 17, 30, 45, 60 & 0$\sim$360 & X \& Ku & quad & 18,071 & 18,071 \\
 & test & \textbf{50} & all & 15, 30, 45, 60 & 0$\sim$360 & X \& Ku & quad & 17,603 & 17,613 \\ \cmidrule(lr){1-10}
\multirow{2}{*}{EOC-Scene} & train & 40 & \textbf{simple} & 15, 30, 45, 60 & 0$\sim$360 & X \& Ku & quad & 19,584 & 19,584 \\
 & test & 40 & \textbf{difficult} & 15, 30, 45, 60 & 0$\sim$360 & X \& Ku & quad & 77,791 & 77,365 \\ \cmidrule(lr){1-10}
\multirow{2}{*}{EOC-Depression} & train & 40 & all & \textbf{15} & 0$\sim$360 & X \& Ku & quad & 24,361 & 22,206 \\
 & test & 40 & all & \textbf{30, 45, 60} & 0$\sim$360 & X \& Ku & quad & 73,014 & 74,743 \\ \cmidrule(lr){1-10}
\multirow{2}{*}{EOC-Azimuth} & train & 40 & all & 15, 30, 45, 60 & \textbf{0$\sim$60} & X \& Ku & quad & 18,636 & 18,592 \\
 & test & 40 & all & 15, 30, 45, 60 & \textbf{60$\sim$360} & X \& Ku & quad & 78,739 & 78,357 \\ \cmidrule(lr){1-10}
\multirow{2}{*}{EOC-Band} & train & 40 & except city & 15, 30, 45, 60 & 0$\sim$360 & \textbf{X} & quad & 27,711 & 27,653 \\
 & test & 40 & except city & 15, 30, 45, 60 & 0$\sim$360 & \textbf{Ku} & quad & 27,763 & 27,732 \\ \cmidrule(lr){1-10}
\multirow{2}{*}{EOC-Polarization} & train & 40 & all & 15, 30, 45, 60 & 0$\sim$360 & X \& Ku & \textbf{HH} & 24,361 & 24,246 \\
 & test & 40 & all & 15, 30, 45, 60 & 0$\sim$360 & X \& Ku & \textbf{other} & 73,014 & 72,703 \\ \bottomrule
\end{tabular}
\end{table*}

\textbf{SOC and EOC settings -} 
SOC settings are those where the training and test sets have similar imaging conditions, and we have SOC-40 and SOC-50 settings. 
\emph{SOC-40} is created from a random data sample with a 7:3 training-to-test ratio. \emph{SOC-50} randomly selects data with a similar amount of MSTAR ten classes, and we combine our dataset with MSTAR. 
EOC settings are those where there is a significant domain shift between the training set and the test set, such as variation in imaging conditions and target state. 
For \emph{EOC-Scene}, we use simple scenes (sandstone and bare soil) as the training set and complex scenes (urban, factory, and woodland) with occlusion as the test set. \emph{EOC-Depression} is the training set with a 15$^{\circ}$ depression angle and the test set with 30$^{\circ}$, 45$^{\circ}$ and 60$^{\circ}$. 
\emph{EOC-Azimuth} means that training has limited target azimuth angles within 0$^{\circ}$$\sim$60$^{\circ}$, and the test has other angles. Imaging angle and geometry variation can change target signatures and background clutter. 
\emph{EOC-band} and \emph{EOC-polarization} are designed to test the effect of different bands and change from HH-polarization to other polarizations on recognition, respectively. We do not consider experimental settings that added simulation perturbations, such as random/Gaussian noise and simulation occlusions.

\textbf{Compared methods -} 
We select classical or recent open-source work in computer vision and SAR target recognition, in addition to recently proposed methods from our laboratory. For classification, we have chosen six computer vision methods (VGG16~\cite{ref32}, ResNet18~\cite{he2016deep}, ResNet34~\cite{he2016deep}, ConvNeXt~\cite{liu2022convnet}, ViT~\cite{dosovitskiy2020image}, and HiViT~\cite{zhang2023hivit}) and four SAR ATR methods (HDANet~\cite{10283916}, SARATR-X~\cite{li2024saratrx,li2024predicting}, MS-CVNets~\cite{zeng2022sar}, and LDSF~\cite{10753051}). MS-CVNets and LDSF are complex images methods others are magnitude methods. VGG16, ResNet18, ResNet34, ConvNeXt, ViT, HiViT, and SARATR-X all use the pre-training weights.
For detection, we have chosen four computer vision methods (Faster RCNN~\cite{ren2016fasterrcnn}, CenterNet~\cite{zhou2019objects}, YOLOv8~\cite{yolov8}, and YOLOv10~\cite{wang2024yolov10}) and two SAR ATR methods (DiffDet4SAR~\cite{zhou2024diffdet4sar}, and SARATR-X~\cite{li2024saratrx,li2024predicting}). These methods are based on magnitude images. Because of the long-term dataset and benchmark construction process, we will keep updating the homepage with new experimental settings and method results. 

\textbf{Transfer learning -} 
\textcolor{black}{To evaluate whether domain-specific pre-training (using our SAR dataset) offers advantages over large-scale natural image pre-training (such as ImageNet)}.
we perform training on SOC-40 according to Table~\ref{table:Classification results}, and the weights are used \textcolor{black}{as initialization for fine-tuning on other SAR target classification dataset}. Fine-tuning parameter settings follow those in our previous paper~\cite{li2024predicting}, and performance is compared with models initialized with ImageNet-pretrained weights.



\subsection{Results and analyses}
Here, we present the results \textcolor{black}{and analyses of the recognition tasks in ATRBench}.

\begin{table*}[!tb]
\centering
\footnotesize
\caption{
\textbf{Classification results.} We use overall accuracy ($\%$) as a metric, \emph{i.e.}, the number of correctly classified samples in proportion to the total number of samples. \textcolor{gray}{Gray color} methods use complex images, and magnitude image methods are sorted by network architecture, such as CNN and Transformer.  \textbf{Bolded} text indicates the best result, while \underline{underlined} text is the next best result.
}
\label{table:Classification results}
\renewcommand\arraystretch{1.1}
\begin{tabular}{ccccccccc|cc} 
\toprule
\makecell[c]{Setting} & \makebox[0.9cm][c]{\fontsize{6}{6}\selectfont\makecell[c]{VGG16\\\cite{ref32}}} & \makebox[0.9cm][c]{\fontsize{6}{6}\selectfont\makecell[c]{HDANet\\\cite{10283916}}} & \makebox[0.9cm][c]{\fontsize{6}{6}\selectfont\makecell[c]{ResNet18\\\cite{he2016deep}}} & \makebox[0.9cm][c]{\fontsize{6}{6}\selectfont\makecell[c]{ResNet34\\\cite{he2016deep}}} & \makebox[0.9cm][c]{\fontsize{6}{6}\selectfont\makecell[c]{ConvNeXt\\\cite{liu2022convnet}}} & \makebox[0.9cm][c]{\fontsize{6}{6}\selectfont\makecell[c]{ViT\\\cite{dosovitskiy2020image}}} & \makebox[0.9cm][c]{\fontsize{6}{6}\selectfont\makecell[c]{HiViT\\\cite{zhang2023hivit}}} & \makebox[0.9cm][c]{\fontsize{6}{6}\selectfont\makecell[c]{SARATR-X\\\cite{li2024saratrx,li2024predicting}}} & \makebox[0.9cm][c]{\fontsize{6}{6}\selectfont\makecell[c]{\textcolor{gray}{MS-CVNet}\\\cite{zeng2022sar}}} & \makebox[0.9cm][c]{\fontsize{6}{6}\selectfont\makecell[c]{\textcolor{gray}{LDSF}\\\cite{10753051}}} \\ 
\cmidrule(lr){1-11}
SOC-40 & 88.8 & 89.1 & 90.6 & 91.7 & \underline{96.0} & 76.4 & 86.8 & \textbf{96.4} & 80.9 & 88.2 \\ 
\cmidrule(lr){1-11}
SOC-50 & 72.9 & 63.7 & 71.2 & 72.9 & \underline{81.6} & 59.2 & 68.0 & \textbf{85.2} & 55.2 & 65.7 \\ 
\cmidrule(lr){1-11}
EOC-Scene & 21.6 & \textbf{33.6} & 16.1 & 18.0 & 16.5 & 12.9 & 15.8 & 19.5 & 26.1 & \underline{29.7} \\
– (city) & 22.7 & \textbf{33.7} & 18.1 & 18.9 & 17.9 & 11.6 & 13.9 & 20.4 & 26.1 & \underline{32.7} \\
– (factory) & 20.5 & \textbf{34.3} & 13.7 & 16.8 & 14.4 & 15.3 & 17.8 & 18.4 & 25.6 & \underline{30.1} \\
– (woodland) & 6.8 & \textbf{27.6} & 4.2 & 5.4 & 4.2 & 1.97 & 3.8 & 2.7 & 20.2 & \underline{26.3} \\ 
\cmidrule(lr){1-11}
EOC-Depression & 33.2 & 32.9 & 33.9 & 37.2 & \textbf{43.1} & 30.4 & 31.4 & \underline{39.9} & 21.8 & 28.3 \\
– (30$^{\circ}$) & 52.4 & 54.0 & 52.2 & 55.9 & \textbf{63.2} & 43.7 & 47.3 & \underline{58.1} & 39.2 & 46.5 \\
– (45$^{\circ}$) & 33.5 & 32.7 & 34.2 & 38.3 & \textbf{45.1} & 31.0 & 32.0 & \underline{41.2} & 19.2 & 25.9 \\
– (60$^{\circ}$) & 14.0 & 12.3 & 15.4 & 17.4 & \textbf{21.3} & 16.7 & 15.0 & \underline{20.5} & 8.1 & 12.7 \\ 
\cmidrule(lr){1-11}
EOC-Azimuth & 15.7 & 16.4 & 14.9 & 16.5 & 21.1 & \textbf{29.0} & 22.8 & \underline{26.4} & 19.0 & 22.5 \\
– (60$^{\circ}$ $\sim$ 120$^{\circ}$) & 20.7 & 19.4 & 18.0 & 22.1 & 26.2 & \textbf{34.4} & 27.2 & \underline{28.9} & 26.8 & 31.0 \\
– (120$^{\circ}$ $\sim$ 180$^{\circ}$) & 8.2 & 9.5 & 8.3 & 8.5 & 13.5 & \textbf{26.4} & 14.2 & \underline{22.8} & 13.2 & 15.6 \\
– (180$^{\circ}$ $\sim$ 240$^{\circ}$) & 17.6 & 17.9 & 16.4 & 17.4 & 19.9 & 20.8 & \underline{22.2} & \textbf{23.0} & 16.9 & 19.7 \\
– (240$^{\circ}$ $\sim$ 300$^{\circ}$) & 7.11 & 7.0 & 7.4 & 7.9 & 10.6 & \textbf{20.3} & 11.1 & \underline{12.3} & 10.9 & 12.6 \\
– (300$^{\circ}$ $\sim$ 360$^{\circ}$) & 26.0 & 30.4 & 25.5 & 27.5 & 36.6 & \underline{44.4} & 40.9 & \textbf{46.7} & 28.7 & 33.7 \\ 
\cmidrule(lr){1-11}
EOC-Band & 78.8 & 79.5 & 83.1 & 83.8 & \underline{88.4} & 65.7 & 70.7 & \textbf{89.2} & 63.7 & 60.9 \\
– inverse & 74.5 & 79.8 & 76.1 & 79.0 & \underline{84.6} & 62.3 & 78.4 & \textbf{89.1} & 60.9 & 64.2 \\ 
\cmidrule(lr){1-11}
EOC-Polarization & 72.5 & 63.1 & 71.4 & 70.5 & \underline{83.1} & 53.6 & 67.1 & \textbf{84.6} & 49.0 & 55.1 \\
– VV & 72.4 & 63.0 & 72.9 & 72.1 & \underline{84.4} & 55.2 & 69.1 & \textbf{87.5} & 52.0 & 63.4 \\
– HV & 72.3 & 62.7 & 70.2 & 69.4 & \underline{82.4} & 52.4 & 65.9 & \textbf{83.1} & 45.8 & 49.0 \\
– VH & 72.8 & 63.5 & 71.0 & 70.0 & \underline{82.7} & 55.1 & 66.5 & \textbf{83.3} & 46.1 & 52.8 \\
\bottomrule
\end{tabular}
\end{table*}

\textbf{Classification -}  
From different methods in Table~\ref{table:Classification results}, we note that deep learning methods can achieve high performance in SAR fine-grained classification, such as the recent methods ConvNeXt and SARATR-X in SOC-40. Methods based on complex images, such as MS-CVNet and LDSF, better distinguish between targets and clutter in EOC-Scene. The model recognition performance and robustness of the latest 2020s CNN architecture ConvNeXt are superior to the previous CNN models. ConvNeXt is more robust under EOC-Deprssion than other methods, and ViT is insensitive to azimuthal angle change. Our previously proposed method, HDANet, improves the attention mechanism and uses a segmentation task so it can be used in different scenarios. In addition, our proposed LDSF introduces a scattering graph structure that effectively improves the discrimination of targets and clutter. SARATR-X is a pre-trained self-supervised learning model on SAR with the HiViT architecture, and its performance is significantly improved compared to HiViT. Besides, the Transformer models are trained with some low-layer weights frozen, as we found that the full update tends to lead to overfitting.

\begin{figure}[!tb]
\centering
\includegraphics[width=8.4cm]{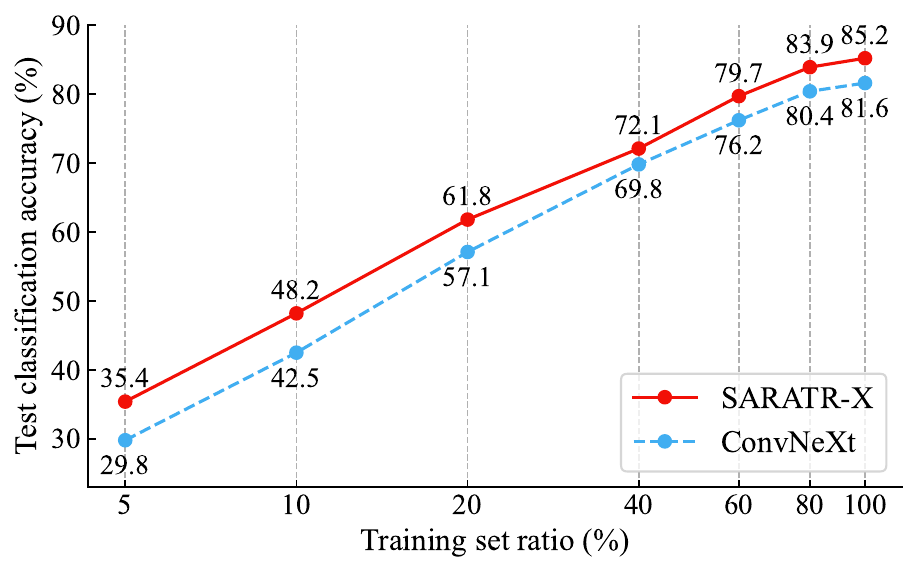}
\caption{
\textbf{Results under SOC-50 with different training set ratio.} 
We find a linear trend decreasing test set accuracy as training decreases on a logarithmic scale.
}
\label{fig_few_shot}
\end{figure}

The different experimental settings in Table~\ref{table:Classification results} show that there are still great challenges in SAR target recognition. As the number of fine-grained classes increased, previous methods were no longer able to achieve nearly 99\% accuracy on ATRNet-STAR's SOC. \textcolor{black}{A deeper challenge lies in the generalization ability of models from simple to complex conditions.} In the difficult scene of EOC-Scene, deep learning encounters great challenges, especially in the background clutter interference inside the trees in the woodland. The significant deterioration in accuracy indicates that deep learning models trained on simple backgrounds have difficulty distinguishing between targets and complex background clutter. The use of complex information or the inclusion of auxiliary tasks such as segmentation during training can enhance the recognition ability, but the alteration of target features due to background clutter interference is still a difficult challenge. 
We observe a significant decrease in the complex data compared to the magnitude data under angles and sensor parameter variations. These results are because the complex phase is more sensitive to these parameters' transformations, and the magnitude of the data is additionally preprocessed to enhance the image's visual quality. Changes in imaging angle have a greater impact on target characteristics than band and polarization, especially large depression and azimuth angle changes. Different imaging angles lead to variations in target and background clutter, as well as different occlusions and scattering resulting from changes in imaging geometry. We have not found a good method to deal with the above scene and angle changes. The shift in target features due to band and polarization can be solved very effectively by ConvNeXt and SARATR-X. 
\textcolor{black}{The impact of training data volume on model performance is another critical factor. The test set accuracy tends to decrease linearly as the training samples are reduced on a logarithmic scale in Fig~\ref{fig_few_shot} under the SOC-50 setting.}
However, due to cost issues, we have not discussed the effect of different platform resolutions on recognition in the current version, which is another important issue.

\textbf{Detection -} 
As illustrated in Table~\ref{table:Detection results}, we find that the two-stage approach (Faster RCNN, DiffDet4SAR, and SARATR-X) has better robustness than the single-stage approach (CenterNet, YOLOv8, and YOLOv10) under most EOC conditions. In particular, CenterNet models the target center using a Gaussian kernel, demonstrating superior detection capabilities under EOC-Depression and EOC-Polarization conditions. However, YOLO series methods, which rely on fixed grid partitioning and coordinate regression, exhibit limitations in detection performance under EOC conditions. In contrast, our proposed DiffDet4SAR, a diffusion model-based detection framework, consistently outperforms other algorithms in EOC-SCENE and EOC-Azimuth scenarios. This advantage stems from the robust denoising box structure, which enhances target location accuracy, and the use of central difference convolution to amplify target saliency. The foundation model SARATR-X, based on self-supervised learning with SAR images, performs well for fine-grained recognition and most of the EOC conditions, but the performance for scene and angle changes needs to be improved. 

\begin{table}[!tb]
\centering
\footnotesize
\caption{
\textbf{Detection results.} We use mAP50~\cite{liu2020deep,Chen2025} as a metric, \emph{i.e.}, the average precision at IoU 0.5 because the horizontal bounding box does not exactly fit the contours of the different targets. \textbf{Bolded} text indicates the best result, while \underline{underlined} text is the next best result.
}
\label{table:Detection results}
\renewcommand\arraystretch{1.1}
\resizebox{0.95\linewidth}{!}{%
\begin{tabular}{ccccccc} 
\toprule
\makecell[c]{Setting} & \makebox[0.9cm][c]{\fontsize{6}{6}\selectfont\makecell[c]{CenterNet\\\cite{zhou2019objects}}} & \makebox[0.9cm][c]{\fontsize{6}{6}\selectfont\makecell[c]{YOLOv8\\\cite{yolov8}}} & \makebox[0.9cm][c]{\fontsize{6}{6}\selectfont\makecell[c]{YOLOv10\\\cite{wang2024yolov10}}} & \makebox[0.9cm][c]{\fontsize{6}{6}\selectfont\makecell[c]{Faster RCNN\\\cite{ren2016fasterrcnn}}} & \makebox[0.9cm][c]{\fontsize{6}{6}\selectfont\makecell[c]{DiffDet4SAR\\\cite{zhou2024diffdet4sar}}} & \makebox[0.9cm][c]{\fontsize{6}{6}\selectfont\makecell[c]{SARATR-X\\\cite{li2024saratrx,li2024predicting}}} \\ 
\cmidrule(lr){1-7}
SOC-40 & 83.5 & 86.1 & \underline{91.7} & 88.3 & 90.9 & \textbf{95.5} \\ 
\cmidrule(l){1-7}
SOC-50 & 74.7 & 64.4 & 63.5 & 65.0 & \underline{76.1} & \textbf{79.6} \\ 
\cmidrule(l){1-7}
EOC-Scene & 30.2 & 26.2 & 21.5 & \underline{35.5} & \textbf{37.2} & 20.4 \\
– (city) & 37.8 & 30.8 & 24.8 & \underline{42.6} & \textbf{45.9} & 18.1 \\
– (factory) & 27.1 & 24.3 & 20.3 & \underline{32.9} & \textbf{35.2} & 24.1 \\
– (woodland) & 23.8 & 25.1 & 20.0 & \underline{30.7} & \textbf{32.8} & 25.2 \\ 
\cmidrule(l){1-7}
EOC-Depression & \textbf{33.5} & 22.7 & 22.3 & 29.0 & 31.0 & \underline{33.3} \\
– (30$^{\circ}$) & \textbf{57.3} & 42.9 & 42.5 & 50.1  & 52.1 & \underline{55.4} \\
– (45$^{\circ}$) & \underline{34.2} & 21.4 & 21.6 & 27.7 & 30.1 & \textbf{35.0} \\
– (60$^{\circ}$) & 8.1 & 4.4 & 4.1 & 6.4 & \underline{8.2} & \textbf{9.9} \\ 
\cmidrule(l){1-7}
EOC-Azimuth & 24.5 & 18.1 & 19.2 & \underline{26.9} & \textbf{28.7} & 22.9 \\
– (60$^{\circ}$ $\sim$ 120$^{\circ}$) & 21.4 & 15.9 & 18.6 & \underline{26.8} & \textbf{28.9} & 20.5 \\
– (120$^{\circ}$ $\sim$ 180$^{\circ}$) & 20.0 & 13.0 & 14.3 & 20.8 & \underline{22.1} & \textbf{22.3} \\
– (180$^{\circ}$ $\sim$ 240$^{\circ}$) & 18.7 & 15.0 & 14.1 & \underline{19.0 } & \textbf{20.1} & 18.6 \\
– (240$^{\circ}$ $\sim$ 300$^{\circ}$) & 14.9 & 10.4 & 11.4 & \underline{18.0} & \textbf{19.0} & 14.0 \\
– (300$^{\circ}$ $\sim$ 360$^{\circ}$) & 56.0 & 44.6 & 44.7 & \underline{57.9} & \textbf{60.1} & 46.7 \\ 
\cmidrule(l){1-7}
EOC-Band & 69.1 & 72.4 & 74.5 & 72.9 & \underline{75.1} & \textbf{87.5} \\
– inverse & \underline{75.9} & 75.2 & 72.1 & 69.5 & 73.8 & \textbf{83.7} \\ 
\cmidrule(l){1-7}
EOC-Polarization & \underline{74.8} & 56.0 & 59.2 & 58.3 & 61.2 & \textbf{75.3} \\
– VV & \underline{75.6} & 56.9 & 62.4 & 57.9 & 60.4 & \textbf{77.4} \\
– HV & \textbf{74.3} & 55.3 & 57.6 & 59.0 & 62.1 & \underline{73.9} \\
– VH & \textbf{75.1} & 56.4 & 58.3 & 58.5 & 61.0 & \underline{74.8} \\
\bottomrule
\end{tabular}
}
\end{table}

Compared with the classification task, the detection task can effectively distinguish between target and background clutter, especially for target discovery in complex scenes such as woodlands. However, most of the detection performance is slightly lower than the classification performance due to the SAR image quality issues that result in noisy labels and rectangular box offsets. Besides, our fine-grained detection tasks, encompassing both accurate localization and precise classification, are inherently more complex than single precise classification tasks, necessitating advanced model architectures and sophisticated algorithm designs. Furthermore, although we changed the target position, such adjustments are not performed frequently due to acquisition time and cost constraints. Similarly, there is still much scope for improvement in robustness under scene and imaging angle variations.

\begin{figure}[!tb]
\centering
\includegraphics[width=8.0cm]{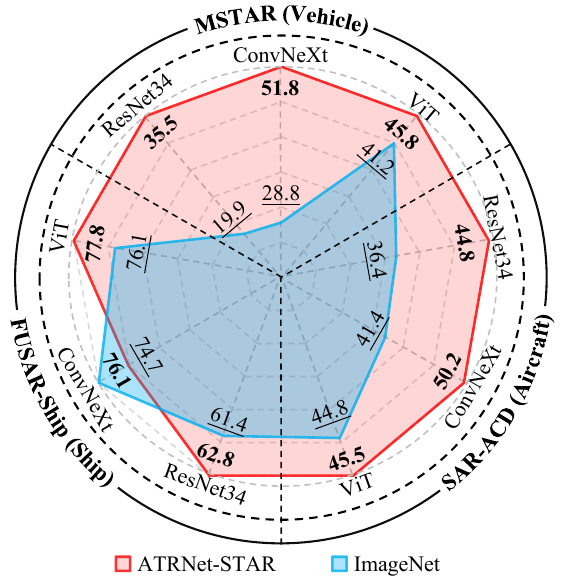}
\caption{
\textbf{Transfer learning results.} 
We report the 5-shot fine-tuning performance of transfer learning on three SAR target classification datasets after ATRNet-STAR training. 
}
\label{fig_transfer}
\end{figure}

\textbf{Transfer learning -}
Fig~\ref{fig_transfer} shows how our dataset can help different models benefit from feature learning and be used for target recognition in other datasets. We use Table's SOC-40 \textcolor{black}{supervised pre-training} weights as initializations on other target datasets with 5-shot sample experiments~\cite{li2024predicting}. The fine-tuning results show that similar ground target-type datasets, such as vehicles and airplanes, would greatly benefit from training on our data. However, the transfer effect is not obvious for the ship target class dataset in the sea scene. In recent years, the number of emerging satellite ship datasets has been much larger than that of other ground targets. This dataset can greatly compensate for the lack of high-resolution ground target samples in the existing SAR target datasets and advance the construction of SAR ATR foundation models.

In conclusion, we evaluate our benchmarks based on 7 experimental settings and 15 recognition methods with classification and detection tasks based on magnitude images and classification tasks based on complex images. These experiments provide valuable insights and substantial potential into recognizing SAR target characteristics under different imaging conditions and fine-grained classes. Our dataset is sufficient to provide a new, challenging benchmark for advancing SAR target recognition technology. We also welcome researchers to contact us to construct richer experimental benchmarks from original data.

\subsection{Discussion}
Researchers have endeavored to realize a stable and efficient SAR ATR system, but previous vehicle benchmark datasets have not provided sufficiently challenging conditions. Our results show that representative methods encounter significant challenges, such as imaging angle and scene variation, as well as few-shot settings. Robust feature extraction under different varying imaging conditions remains a critical research problem. It is also crucial to efficiently utilize a small number of labeled samples and mine complex and polarization information. Although the pre-trained foundation model SARATR-X is effective in most cases compared to other methods, no single method can effectively handle diverse perturbation conditions, highlighting the complexity of SAR ATR. In a data-driven approach dominant in the 2020s, our transfer learning results show that a large-scale dataset will help to extract stabilizing features and enhance performance across multiple datasets and tasks. Based on the results, we discuss several potential directions for the future development of ATR systems.

\textbf{Robustness under different conditions -} 
Imaging condition variations present a significant challenge for recognition in Table~\ref{table:SOC and EOC settings}. This variability in the domain distribution requires the extraction of stable invariant target features across conditions.

\textbf{Efficient utilization of samples -} 
Given the challenges associated with acquiring and labeling SAR images, it is unlikely that the sample size will reach the same size as the visible-light dataset. Therefore, it is important to improve learning efficiency by utilizing a small number of samples in Fig~\ref{fig_few_shot}.

\textbf{Incorporation with multi-dimensional information -}   
Table~\ref{table:Classification results} show that complex information can help distinguish between targets and clutter under scene variations. Auxiliary information such as complex and angle can be introduced in SAR ATR, but increasing the dimensionality of the information may increase the instability and complexity.

\textbf{Fast incremental learning capability -} 
Because of the open real world, ATR systems face constantly changing imaging conditions and target types. They need to effectively recognize anomalies and update their capabilities in response to variations. This dataset provides a rich set of target types and imaging conditions that can be used to explore this possibility.

\textbf{Explainable, interpretable, and trustworthy recognition model -}
Although data-driven models have become a mainstream approach for SAR ATR, the requirement for black-box models' interpretability during the recognition process makes it difficult to meet the trust needed by actual systems. Therefore, in addition to improving recognition performance, a transparent and trustworthy recognition model is also a problem that needs to be studied~\cite{huang2022Progress, li2022research, datcu2023explainable}.

Subsequently, we plan to construct a different resolution version and paired polarization data. We will continue to expand the scenes and target classes by combining satellite data with a semi-automated and low-cost method.

\section{Conclusion}
\label{sec:Conclusion}
A large-scale fine-grained SAR vehicle dataset benchmark, named ATRNet-STAR, has been established in this paper. It contains various target types, scene variations, and imaging conditions and is 10 times the size of previous datasets of the same type. We provide a comprehensive benchmark ATRBench of 7 experiment settings and 15 methods. It will facilitate diverse explorations of issues and promote SAR ATR techniques. In the future, we will continue to contribute to the SAR target datasets, and we welcome researchers to contact us to promote the progress of the SAR ATR field jointly.

\footnotesize
\bibliographystyle{IEEEtran}
\bibliography{IEEEabrv,ref}

\end{document}